\documentclass[10pt,twocolumn,letterpaper]{article}

\usepackage{cvpr}              

\usepackage{graphicx}
\usepackage{amsmath}
\usepackage{amssymb}
\usepackage{booktabs}
\usepackage{amsmath}
\usepackage{amssymb}
\usepackage{comment}
\usepackage{bm}
\usepackage{bbm}
\usepackage{algorithm}
\usepackage{algpseudocode}
\usepackage{mathtools}
\usepackage{multirow}
\usepackage{multicol}
\DeclarePairedDelimiterX{\infdivx}[2]{(}{)}{%
  #1\;\delimsize|\delimsize|\;#2%
}
\newcommand{\kld}[2]{\ensuremath{D_{KL}\infdivx{#1}{#2}}\xspace}
\DeclareMathOperator*{\argmin}{argmin}

\usepackage[pagebackref,breaklinks,colorlinks]{hyperref}

\newlength\savewidth\newcommand\shline{\noalign{\global\savewidth\arrayrulewidth
  \global\arrayrulewidth 1pt}\hline\noalign{\global\arrayrulewidth\savewidth}}

\usepackage[capitalize]{cleveref}
\crefname{section}{Sec.}{Secs.}
\Crefname{section}{Section}{Sections}
\Crefname{table}{Table}{Tables}
\crefname{table}{Tab.}{Tabs.}

\def\cvprPaperID{3874} 
\def\confName{CVPR}
\def\confYear{2022}

\begin{document}

\title{Vector Quantized Diffusion Model for Text-to-Image Synthesis}

\author{Shuyang Gu$^{1}$ \qquad Dong Chen$^{2}$ \qquad Jianmin Bao$^{2}$ \qquad Fang Wen$^{2}$ \vspace{1pt}\\
Bo Zhang$^{2}$ \qquad Dongdong Chen$^{3}$ \qquad Lu Yuan$^{3}$\qquad Baining Guo$^{2}$  \vspace{1pt}\\
$^{1}$University of Science and Technology of China\qquad$^{2}$Microsoft Research\qquad$^{3}$Microsoft Cloud+AI\\
\hspace{0.1in}{\tt\small gsy777@mail.ustc.edu.cn} \quad {\tt\small \{doch,jianbao,fangwen,zhanbo,dochen,luyuan,bainguo\}@microsoft.com} \\
}
\maketitle

\begin{abstract}

We present the vector quantized diffusion (VQ-Diffusion) model for text-to-image generation. This method is based on a vector quantized variational autoencoder (VQ-VAE) whose latent space is modeled by a conditional variant of the recently developed Denoising Diffusion Probabilistic Model (DDPM). We find that this latent-space method is well-suited for text-to-image generation tasks because it not only eliminates the unidirectional bias with existing methods but also allows us to incorporate a mask-and-replace diffusion strategy to avoid the accumulation of errors, which is a serious problem with existing methods. Our experiments show that the VQ-Diffusion produces significantly better text-to-image generation results when compared with conventional autoregressive (AR) models with similar numbers of parameters. Compared with previous GAN-based text-to-image methods, our VQ-Diffusion can handle more complex scenes and improve the synthesized image quality by a large margin. Finally, we show that the image generation computation in our method can be made highly efficient by reparameterization. With traditional AR methods, the text-to-image generation time increases linearly with the output image resolution and hence is quite time consuming even for normal size images. The VQ-Diffusion allows us to achieve a better trade-off between quality and speed. Our experiments indicate that the VQ-Diffusion model with the reparameterization is fifteen times faster than traditional AR methods while achieving a better image quality. The code and models are available at \url{https://github.com/cientgu/VQ-Diffusion}.

\end{abstract}

\section{Introduction}

Recent success of Transformer~\cite{vaswani2017attention, devlin2018bert} in neural language processing (NLP) has raised tremendous interest in using successful language models for computer vision tasks. Autoregressive (AR) model~\cite{radford2018improving,radford2019language,brown2020language} is one of the most natural and popular approach to transfer from text-to-text generation (\ie, machine translation) to text-to-image generation. Based on the AR model, recent work DALL-E~\cite{ramesh2021zero} has achieved impressive results for text-to-image generation.

Despite their success, existing text-to-image generation methods still have weaknesses that need to be improved. One issue is the unidirectional bias. Existing methods predict pixels or tokens in the reading order, from top-left to bottom-right, based on the attention to all prefix pixels/tokens and the text description. This fixed order introduces unnatural bias in the synthesized images because important contextual information may come from any part of the image, not just from left or above. Another issue is the accumulated prediction errors. Each step of the inference stage is performed based on previously sampled tokens – this is different from that of the training stage, which relies on the so-called “teacher-forcing” practice~\cite{esser2021imagebart} and provides the ground truth for each step. This difference is important and its consequence merits careful examination. In particular, a token in the inference stage, once predicted, cannot be corrected and its errors will propagate to the subsequent tokens.

We present the vector quantized diffusion (VQ-Diffusion) model for text-to-image generation, a model that eliminates the unidirectional bias and avoids accumulated prediction errors. We start with a vector quantized variational autoencoder (VQ-VAE) and model its latent space by learning a parametric model using a conditional variant of the Denoising Diffusion Probabilistic Model (DDPM)~\cite{sohl2015deep,ho2020denoising}, which has been applied to image synthesis with compelling results~\cite{dhariwal2021diffusion}. We show that the latent-space model is well-suited for the task of text-to-image generation. Roughly speaking, the VQ-Diffusion model samples the data distribution by reversing a forward diffusion process that gradually corrupts the input via a fixed Markov chain. The forward process yields a sequence of increasingly noisy latent variables of the same dimensionality as the input, producing pure noise after a fixed number of timesteps. Starting from this noise result, the reverse process gradually denoises the latent variables towards the desired data distribution by learning the conditional transit distribution.

The VQ-Diffusion model eliminates the unidirectional bias. It consists of an independent text encoder and a diffusion image decoder, which performs denoising diffusion on discrete image tokens. At the beginning of the inference stage, all image tokens are either masked or random. Here the masked token serves the same function as those in mask-based generative models~\cite{devlin2018bert}. The denoising diffusion process gradually estimates the probability density of image tokens step-by-step based on the input text. In each step, the diffusion image decoder leverages the contextual information of all tokens of the entire image predicted in the previous step to estimate a new probability density distribution and use this distribution to predict the tokens in the current step. This bidirectional attention provides global context for each token prediction and eliminates the unidirectional bias.

The VQ-Diffusion model, with its mask-and-replace diffusion strategy, also avoids the accumulation of errors. In the training stage, we do not use the “teacher-forcing” strategy. Instead, we deliberately introduce both masked tokens and random tokens and let the network learn to predict the masked token and modify incorrect tokens. In the inference stage, we update the density distribution of all tokens in each step and resample all tokens according to the new distribution. Thus we can modify the wrong tokens and prevent error accumulation. Comparing to the conventional replace-only diffusion strategy for unconditional image generation~\cite{austin2021structured}, the masked tokens effectively direct the network’s attention to the masked areas and thus greatly reduce the number of token combinations to be examined by the network. This mask-and-replace diffusion strategy significantly accelerates the convergence of the network.

To assess the performance of the VQ-Diffusion method, we conduct text-to-image generation experiments with a wide variety of datasets, including CUB-200~\cite{wah2011caltech}, Oxford-102~\cite{nilsback2008automated}, and MSCOCO~\cite{lin2014microsoft}. Compared with AR model with similar numbers of model parameters, our method achieves significantly better results, as measured by both image quality metrics and visual examination, and is much faster. Compared with previous GAN-based text-to-image methods~\cite{xu2018attngan,zhang2017stackgan,zhang2018stackgan++,zhu2019dm}, our method can handle more complex scenes and the synthesized image quality is improved by a large margin. Compared with extremely large models (models with ten times more parameters than ours), including DALL-E~\cite{ramesh2021zero} and CogView~\cite{ding2021cogview}, our model achieves comparable or better results for specific types of images, \ie, the types of images that our model has seen during the training stage. Furthermore, our method is general and produces strong results in our experiments on both unconditional and conditional image generation with FFHQ~\cite{karras2019style} and ImageNet~\cite{deng2009imagenet} datasets.

The VQ-Diffusion model also provides important benefits for the inference speed. With traditional AR methods, the inference time increases linearly with the output image resolution and the image generation is quite time consuming even for normal-size images (\eg, images larger than small thumbnail images of $64 \times 64$ pixels). The VQ-Diffusion provides the global context for each token prediction and makes it independent of the image resolution. This allows us to provide an effective way to achieve a better tradeoff between the inference speed and the image quality by a simple reparameterization of the diffusion image decoder. Specifically, in each step, we ask the decoder to predict the original noise-free image instead of the noise-reduced image in the next denoising diffusion step. Through experiments we have found that the VQ-Diffusion method with reparameterization can be fifteen times faster than AR methods while achieving a better image quality.

\section{Related Work}

\noindent \textbf{GAN-based Text-to-image generation.} 
In the past few years, Generative Adversarial Networks (GANs)~\cite{goodfellow2014generative} have shown promising results on many tasks~\cite{gu2019mask,gu2020giqa,gu2020priorgan},, especially text-to-image generation~\cite{reed2016generative,zhang2017stackgan,dash2017tac,nguyen2017plug,sharma2018chatpainter,hong2018inferring,xu2018attngan,zhang2018stackgan++,zhang2018photographic,gao2019perceptual,lao2019dual,li2019object,qiao2019learn,qiao2019mirrorgan,yin2019semantics,tan2019semantics,zhu2019dm,li2019controllable,cha2019adversarial,el2019tell,cheng2020rifegan,souza2020efficient,liang2020cpgan,tao2020df,zhang2021cross,ruan2021dae,tan2020kt,huang2021unifying}. 
GAN-INT-CLS~\cite{reed2016generative} was the first to use a conditional GAN formulation for text-to-image generation. 
Based on this formulation, some approaches~\cite{zhang2017stackgan,zhang2018stackgan++,xu2018attngan,zhu2019dm,qiao2019mirrorgan,yin2019semantics,zhang2021cross,liang2020cpgan} were proposed to further improve the generation quality.
These models generate high fidelity images on single domain datasets, \eg, birds~\cite{wah2011caltech} and flowers~\cite{nilsback2008automated}. However, due to the inductive bias on the locality of convolutional neural networks, they struggle on complex scenes with multiple objects, such as those in the MS-COCO dataset~\cite{lin2014microsoft}. 

Other works~\cite{hong2018inferring,li2019object} adopt a two-step process which first infer the semantic layout then generate different objects, but this kind of method requires fine-grained object labels, \eg, object bounding boxes or segmentation maps. 

\noindent \textbf{Autoregressive Models.} 
AR models~\cite{radford2018improving,radford2019language,brown2020language} have shown powerful capability of density estimation and have been applied for image generation~\cite{salimans2017pixelcnn++,van2016pixel,parmar2018image,oord2017neural,razavi2019generating,chen2020generative,esser2021taming} recently. PixelRNN~\cite{salimans2017pixelcnn++,van2016pixel}, Image Transformer
~\cite{parmar2018image} and ImageGPT~\cite{chen2020generative} factorized the probability density on an image over raw pixels. Thus, they only generate low-resolution images, like $64 \times 64$, due to the unaffordable amount of computation for large images. 

VQ-VAE~\cite{oord2017neural,razavi2019generating}, VQGAN~\cite{esser2021taming} and ImageBART~\cite{esser2021imagebart} train an encoder to compress the image into a low-dimensional discrete latent space and fit the density of the hidden variables. It greatly improves the performance of image generation. 

DALL-E~\cite{ramesh2021zero}, CogView~\cite{ding2021cogview} and M6~\cite{lin2021m6} propose AR-based text-to-image frameworks. They model the joint distribution of text and image tokens. With powerful large transformer structure and massive text-image pairs, they greatly advance the quality of text-to-image generation, but still have weaknesses of unidirectional bias and accumulated prediction errors due to the limitation of AR models.

\noindent \textbf{Denoising Diffusion Probabilistic Models.}
Diffusion generative models were first proposed in~\cite{sohl2015deep} and achieved strong results on image generation ~\cite{ho2020denoising,nichol2021improved,ho2021cascaded,dhariwal2021diffusion} and image super super-resolution~\cite{saharia2021image} recently.
However, most previous works only considered continuous diffusion models on the raw image pixels. Discrete diffusion models were also first described in~\cite{sohl2015deep}, and then applied to text generation in Argmax Flow~\cite{hoogeboom2021argmax}. D3PMs~\cite{austin2021structured} applies discrete diffusion to image generation. However, it also estimates the density of raw image pixels and can only generate low-resolution (e.g.,$32 \times 32$) images.

\section{Background: Learning Discrete Latent Space of Images Via VQ-VAE}
Transformer architectures have shown great promise in image synthesis due to their outstanding expressivity~\cite{chen2020generative,esser2021taming,ramesh2021zero}. In this work, we aim to leverage the transformer to learn the mapping from text to image. Since the computation cost is quadratic to the sequence length, it is computationally prohibitive to directly model raw pixels using transformers. To address this issue, recent works~\cite{oord2017neural,esser2021taming} propose to represent an image by discrete image tokens with reduced sequence length. Hereafter a transformer can be effectively trained upon this reduced context length and learn the translation from the text to image tokens. 

Formally, a \emph{vector quantized variational autoencoder} (VQ-VAE)~\cite{oord2017neural} is employed. The model consists of an encoder $E$, a decoder $G$ and a codebook $\mathcal{Z} = \{\bm{z}_k\}_{k=1}^K \in \mathbb{R}^{K \times d}$ containing a finite number of embedding vectors , where $K$ is the size of the codebook and $d$ is the dimension of codes. Given an image $\bm{x}\in \mathbb{R}^{H\times W \times 3}$, we obtain a spatial collection of image tokens $\bm{z}_q$ with the encoder  $\bm{z} = E(\bm{x}) \in \mathbb{R}^{h\times w \times d}$ and a subsequent spatial-wise quantizer $Q(\cdot)$ which maps each spatial feature $\bm{z}_{ij}$ into its closest codebook entry $\bm{z}_k$:
\begin{equation}
\vspace{-0.2cm}
	\bm{z}_q = Q(\bm{z}) = \left( \argmin_{\bm{z}_k \in \mathcal{Z}} \|\bm{z}_{ij} - \bm{z}_k\|_2^2 \right) \in \mathbb{R}^{h\times w \times d}
\end{equation}
Where $h \times w$ represents the encoded sequence length and is usually much smaller than $H \times W$. Then the image can be faithfully reconstructed via the decoder, \ie, $\tilde{\bm{x}} = G(\bm{z}_q)$. Hence, image synthesis is equivalent to sampling image tokens from the latent distribution. Note that the image tokens are quantized latent variables in the sense that they take discrete values.  The encoder $E$, the decoder $G$ and the codebook $\mathcal{Z}$ can be trained end-to-end via the following loss function:
\begin{equation}
	\label{eqn:background_L}
	\begin{split}
	\mathcal{L}_\mathrm{VQVAE} = \| \bm{x} - \tilde{\bm{x}} \|_1 + & \| \text{sg}[E(\bm{x})] - {z_q} \|_2^2  \\
	+ \beta & \|\text{sg}[z_q] - E(\bm{x})\|_2^2.
	\end{split}
\end{equation}
Where, $\text{sg}[\cdot]$ stands for the stop-gradient operation. In practice, we replace the second term of  Equation~\ref{eqn:background_L} with exponential moving averages (EMA)~\cite{oord2017neural} to update the codebook entries which is proven to work better than directly using the loss function.

\section{Vector Quantized Diffusion Model}
Given the text-image pairs, we obtain the discrete image tokens $\bm{x} \in \mathbb{Z}^N$ with a pretrained VQ-VAE, where $N=hw$ represents the sequence length of tokens.  Suppose the size of the VQ-VAE codebook is $K$, the image token $x^i$ at location $i$ takes the index that specifies the entries in the codebook, \ie, $x^i\in \{1,2,...,K\}$. On the other hand, the text tokens~$\bm{y} \in \mathbb{Z}^M$ can be obtained through BPE-encoding~\cite{sennrich2015neural}. The overall text-to-image framework can be viewed as maximizing the conditional transition distribution $q(\bm{x}|\bm{y})$.

Previous autoregressive models, \eg, DALL-E~\cite{ramesh2021zero} and CogView~\cite{ding2021cogview}, sequentially predict each image token depends on the text tokens as well as the previously predicted image tokens, \ie, $q(\bm{x}|\bm{y}) = \prod_{i=1}^N q({x}^i|x^1,\cdots,x^{i-1},\bm{y})$. While achieving remarkable quality in text-to-image synthesis, there exist several limitations of autoregressive modeling. First, image tokens are predicted in a unidirectional ordering, \eg, raster scan, which neglects the structure of 2D data and restricts the expressivity for image modeling since the prediction of a specific location should not merely attend to the context on the left or the above. Second, there is a train-test discrepancy as the training employs ground truth whereas the inference relies on the prediction as previous tokens. The so-called “teacher-forcing” practice~\cite{esser2021imagebart} or exposure bias~\cite{schmidt2019generalization} leads to error accumulation due to the mistakes in the earlier sampling. Moreover, it requires a forward pass of the network to predict each token, which consumes an inordinate amount of time even for the sampling in the latent space of low resolution (\ie, $32\times 32$), making the AR model impractical for real usage. 

We aim to model the VQ-VAE latent space in a \emph{non-autoregressive} manner. The proposed VQ-Diffusion method maximizes the probability $q(\bm{x}|\bm{y})$ with the diffusion model~\cite{sohl2015deep,ho2020denoising}, an emerging approach that  produces compelling quality on image synthesis~\cite{dhariwal2021diffusion}. While the majority of recent works focus on continuous diffusion models, using them for categorical distribution is much less researched~\cite{hoogeboom2021argmax,austin2021structured}.  In this work, we propose to use its conditional variant discrete diffusion process for text-to-image generation. We will subsequently introduce the discrete diffusion process inspired by the masked language modeling (MLM)~\cite{devlin2018bert}, and then discuss how to train a neural network to reverse this process. 

\begin{figure}[t]
	\centering
	\vspace{-0.1cm}
	\includegraphics[width=1.0\linewidth]{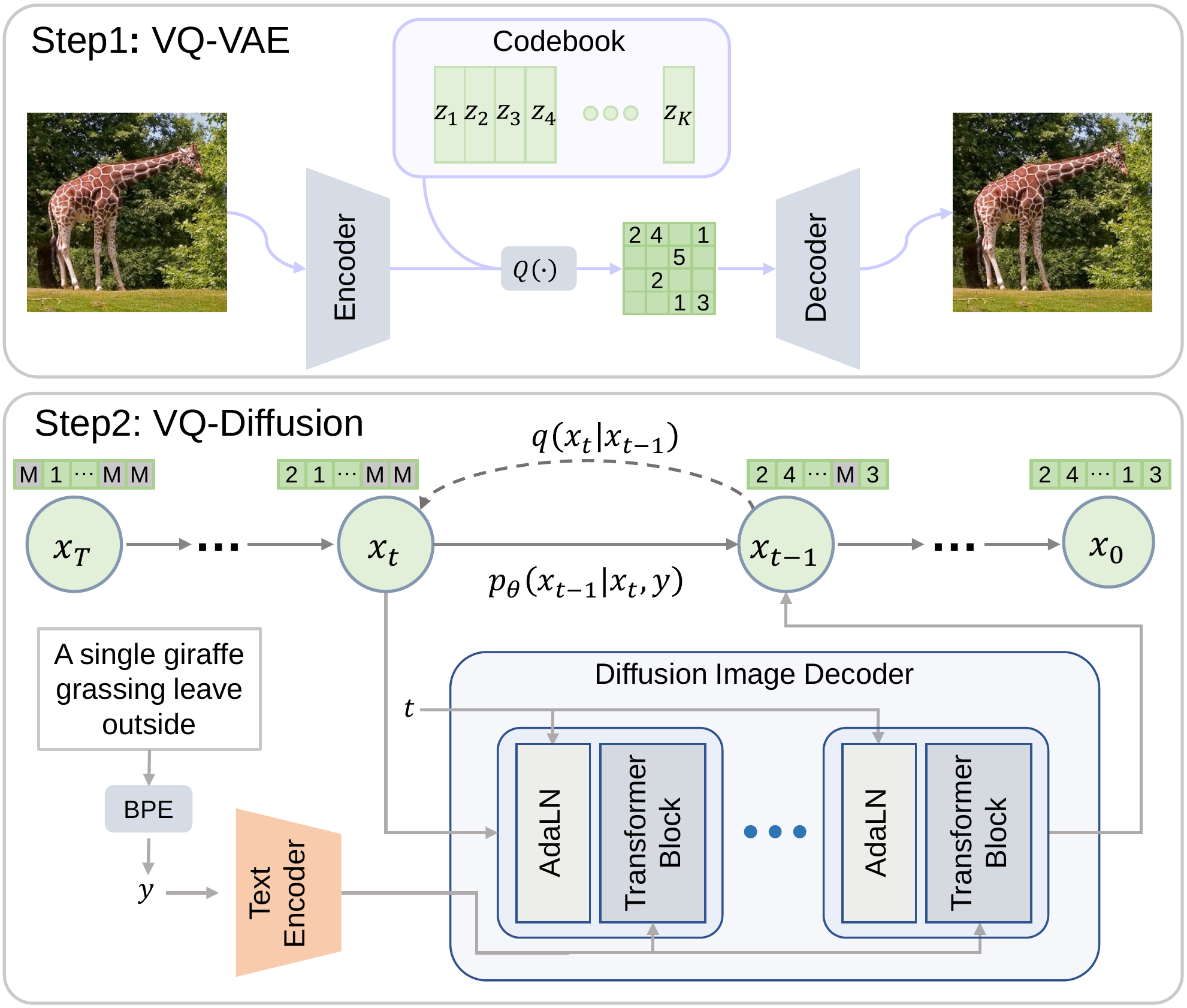}
	\vspace{-0.6cm}
	\caption{Overall framework of our method. It starts with the VQ-VAE. Then, the VQ-Diffusion models the discrete latent space by reversing a forward diffusion process that gradually corrupts the input via a fixed Markov chain.}
	\label{fig:method}
	\vspace{-0.5cm}
\end{figure}

\subsection{Discrete diffusion process}

On a high level,  the \emph{forward} diffusion process gradually corrupts the image data $\bm{x}_0$ via a fixed Markov chain $q(\bm{x}_{t}|\bm{x}_{t-1})$, \eg, random replace some tokens of $\bm{x}_{t-1}$.  After a fixed number of $T$ timesteps, the forward process yields a sequence of increasingly noisy latent variables $\bm{z}_1,...,\bm{z}_T$ of the same dimensionality as $\bm{z}_0$, and $\bm{z}_T$ becomes pure noise tokens. Starting from the noise $\bm{z}_T$, the \emph{reverse} process gradually denoises the latent variables and restore the real data $\bm{x}_0$ by sampling from the reverse distribution $q(\bm{x}_{t-1}|\bm{x}_t, \bm{x}_0)$ sequentially. However, since $\bm{x}_0$ is unknown in the inference stage, we train a transformer network to approximate the conditional transit distribution $p_\theta(\bm{x}_{t-1}|\bm{x}_t, \bm{y})$ depends on the entire data distribution.

To be more specific, consider a single image token $x^i_0$ of $\bm{x}_0$ at location $i$, which takes the index that specifies the entries in the codebook, \ie, $x^i_0\in \{1,2,...,K\}$. Without introducing confusion, we omit superscripts $i$ in the following description. We define the probabilities that $x_{t-1}$ transits to $x_t$ using the matrices $[\bm{Q}_t]_{mn}=q(x_t=m|x_{t-1}=n) \in \mathbb{R}^{K\times K}$. Then the \emph{forward} Markov diffusion process for the whole token sequence can be written as,
\begin{equation}
\vspace{-0.2cm}
q(x_t|x_{t-1}) = \bm{v}^\top(x_{t})\bm{Q}_t \bm{v}(x_{t-1})
\label{eq:markov}
\vspace{-0.02cm}
\end{equation}
where $\bm{v}(x)$ is a one-hot column vector which length is $K$ and only the entry $x$ is 1.  The categorical distribution over $x_t$ is given by the vector $\bm{Q}_t \bm{v}(x_{t-1})$. 

Importantly, due to the property of Markov chain, one can marginalize out the intermediate steps and derive the probability of $x_t$ at arbitrary timestep directly from $x_0$ as,
\begin{equation}
q(x_t|x_0) = \bm{v}^\top(x_{t}) \overline{\bm{Q}}_t \bm{v}(x_0),\  \text{with}\ \overline{\bm{Q}}_t=\bm{Q}_t \cdots \bm{Q}_1.   
\label{eq:forward_chain}
\end{equation}

Besides, another notable characteristic is that by conditioning on $\bm{z}_0$, the posterior of this diffusion process  is tractable, \ie,
\begin{equation}
\vspace{-0.1cm}
\begin{split}
q(x_{t-1}|x_t,x_0) = \frac{q(x_t|x_{t-1},x_0)q(x_{t-1}|x_0)}{q(x_t|x_0)} \\ 
= \frac{
	\left(\bm{v}^\top(x_{t}) {\bm{Q}}_t \bm{v}(x_{t-1})\right) 
	\left(\bm{v}^\top(x_{t-1}) \overline{\bm{Q}}_{t-1} \bm{v}(x_0)\right)}
	{\bm{v}^\top(x_{t}) \overline{\bm{Q}}_t \bm{v}(x_0)}.
\label{eq:posterior}
\end{split}
\vspace{-0.1cm}
\end{equation}

The transition matrix $\bm{Q}_t$ is crucial to the discrete diffusion model and should be carefully designed such that it is not too difficult for the reverse network to recover the signal from noises.

Previous works~\cite{hoogeboom2021argmax,austin2021structured} propose to introduce a small amount of uniform noises to the categorical distribution and the transition matrix can be formulated as,
\begin{equation}
\bm{Q}_t =
\begin{bmatrix}
\alpha_t + \beta_t &  \beta_t & \cdots &\beta_t\\
\beta_t & \alpha_t +  \beta_t & \cdots&\beta_t\\
\vdots & \vdots &  \ddots   & \vdots\\
\beta_t & \beta_t &  \cdots &\alpha_t+\beta_t
\end{bmatrix}
\end{equation}
with $\alpha_t\in[0,1]$ and $\beta_t = (1-\alpha_t)/K$. Each token has a probability of $(\alpha_t+\beta_t)$ to remain the previous value at the current step while with a probability of $K\beta_t$ to be resampled uniformly over all the $K$ categories. 

Nonetheless, the data corruption using uniform diffusion is a somewhat aggressive process that may pose challenge for the reverse estimation. First, as opposed to the Gaussian diffusion process for ordinal data, an image token may be replaced to an utterly uncorrelated category, which leads to an abrupt semantic change for that token. Second, the network has to take extra efforts to figure out the tokens that have been replaced prior to fixing them. In fact, due to the semantic conflict within the local context, the reverse estimation for different image tokens may form a competition and run into the dilemma of identifying the reliable tokens. 

\noindent\textbf{Mask-and-replace diffusion strategy.} To solve the above issues of uniform diffusion, we draw inspiration from mask language modeling~\cite{devlin2018bert} and propose to corrupt the tokens by stochastically masking some of them so that the corrupted locations can be explicitly known by the reverse network. Specifically, we introduce an additional special token, $[\text{\tt{MASK}}]$ token, so each token now has $(K+1)$ discrete states. We define the mask diffusion as follows: each ordinary token has a probability of $\gamma_t$ to be replaced by the $[\text{\tt{MASK}}]$ token and has a chance of $K\beta_t$ to be uniformly diffused, leaving the probability of $\alpha_t = 1-K\beta_t-\gamma_t$ to be unchanged, whereas the $[\text{\tt{MASK}}]$ token always keeps its own state. Hence, we can formulate the  transition matrix $\bm{Q}_t\in \mathbb{R}^{(K+1)\times(K+1)}$ as,
\begin{equation}
   \bm{Q}_t =
   \begin{bmatrix}
   \alpha_t + \beta_t & \beta_t & \beta_t & \cdots & 0 \\
   \beta_t & \alpha_t + \beta_t & \beta_t & \cdots & 0 \\
   \beta_t & \beta_t & \alpha_t + \beta_t & \cdots & 0 \\
   \vdots & \vdots & \vdots & \ddots & \vdots \\
   \gamma_t & \gamma_t &\gamma_t  & \cdots & 1 \\
   \end{bmatrix}.
\label{eq:mask_transit}
\end{equation}

The benefit of this mask-and-replace transition is that: 1) the corrupted tokens are distinguishable to the network, which eases the reverse process. 2) Comparing to the mask only approach in~\cite{austin2021structured}, we theoretically prove that it is necessary to include a small amount of uniform noises besides the token masking, otherwise we get a trivial posterior when $x_t \neq x_0$. 3) The random token replacement forces the network to understand the context rather than only focusing on the $[\text{\tt{MASK}}]$ tokens. 4) The cumulative transition matrix $\overline{\bm{Q}}_t$ and the probability $q(x_t|x_0)$ in Equation~\ref{eq:forward_chain} can be computed in closed form with:
\begin{equation}
	\overline{\bm{Q}}_t \bm{v}(x_0)=\overline{\alpha}_t \bm{v}(x_0) + (\overline{\gamma}_t - \overline{\beta}_t)\bm{v}(K+1) + \overline{\beta}_t
	\label{eqn:fast_Qtv}
\end{equation}
Where $\overline{\alpha}_t=\prod_{i=1}^t\alpha_i$, $\overline{\gamma}_t=1-\prod_{i=1}^t(1-\gamma_i)$, and $\overline{\beta}_t = (1-\overline{\alpha}_t-\overline{\gamma}_t)/K$ can be calculated and stored in advance. Thus, the computation cost of $q(x_t|x_0)$ is reduced from $O(tK^2)$ to $O(K)$. The proof is given in the supplemental material.

\begin{algorithm}[t]
	\caption{Training of the VQ-Diffusion, given transition matrix $\{\bm{Q}_t\}$, initial network parameters $\theta$, loss weight $\lambda$, learning rate $\eta$.} \label{alg:training}
	\begin{algorithmic}[1]
		\Repeat
		\State $(I, s) \gets $ sample training image-text pair 
		\State $\bm{x}_0 \gets \text{VQVAE-Encoder}(I), ~ \bm{y} \gets \text{BPE}(s)$
		\State $t \sim \text{Uniform}(\{1, \cdots, T\})$
		\State $\bm{x}_{t} \gets  \text{sample from}~q(\bm{x}_t|\bm{x}_0)$ \Comment{Eqn.~\ref{eq:forward_chain} and~\ref{eqn:fast_Qtv}}
		\State $\mathcal{L} \gets \begin{cases}
			\mathcal{L}_0, &\text{if } t = 1 \\
			\mathcal{L}_{t-1} + \lambda \mathcal{L}_{x_0}, &\text{otherwise}
		\end{cases}$ \Comment{Eqn.~\ref{eqn:L_vlb} and~\ref{eqn:L_x_0}}
		\State $\theta \gets \theta - \eta \nabla_\theta \mathcal{L}$ \Comment{Update network parameters}
		\Until{converged}
	\end{algorithmic}
\end{algorithm}

\subsection{Learning the reverse process}
To reverse the diffusion process, we train a denoising network $p_\theta(\bm{x}_{t-1}|\bm{x}_t,\bm{y})$ to estimate the posterior transition distribution $q(\bm{x}_{t-1}|\bm{x}_t, \bm{x}_0)$. The network is trained to minimize the variational lower bound (VLB)~\cite{sohl2015deep}:
\begin{equation}
\begin{split}
\mathcal{L}_{\mathrm{vlb}} &= \mathcal{L}_0 + \mathcal{L}_1 + \cdots + \mathcal{L}_{T-1} + \mathcal{L}_{T}, \\ 
\mathcal{L}_0 &= -\log p_\theta(\bm{x}_0|\bm{x}_1,\bm{y}), \\ 
\mathcal{L}_{t-1} &= \kld{q(\bm{x}_{t-1}|\bm{x}_t,\bm{x}_0)}{p_\theta(\bm{x}_{t-1}|\bm{x}_t,\bm{y})}, \\
\mathcal{L}_T &= \kld{q(\bm{x}_T|\bm{x}_0)}{p(\bm{x}_T)}.
\end{split}
\label{eqn:L_vlb}
\end{equation}
Where $p(\bm{x}_T)$ is the prior distribution of timestep $T$. For the proposed mask-and-replace diffusion, the prior is:
\begin{equation}
	\label{eqn:prior_p_x_T}
	p(\bm{x}_T) = \left[\overline{\beta}_T, \overline{\beta}_T, \cdots, \overline{\beta}_T, \overline{\gamma}_T\right]^\top
\end{equation}
Note that since the transition matrix $\bm{Q}_t$ is fixed in the training, the $\mathcal{L}_T$ is a constant number which measures the gap between the training and inference and can be ignored in the training. 

\noindent\textbf{Reparameterization trick on discrete stage.}
The network parameterization affects the synthesis quality significantly. Instead of directly predicting the posterior $q(\bm{x}_{t-1}|\bm{x}_t, \bm{x}_0)$, recent works~\cite{ho2020denoising,hoogeboom2021argmax,austin2021structured} find that approximating some surrogate  variables, \eg, the noiseless target data $q(\bm{x}_0)$ gives better quality. In the discrete setting, we let the network predict the noiseless token distribution $p_\theta(\tilde{\bm{x}}_0 | \bm{x}_t,\bm{y})$ at each reverse step. We can thus compute the reverse transition distribution according to:
\begin{equation}
\vspace{-0.2cm}
\!\!\!\! p_\theta(\bm{x}_{t-1}|\bm{x}_t,\bm{y}) = \sum_{\tilde{\bm{x}}_0=1}^{K} q(\bm{x}_{t-1}|\bm{x}_t,\tilde{\bm{x}}_0) p_\theta(\tilde{\bm{x}}_0|\bm{x}_t, \bm{y}).
\label{eqn:Reparameter_trick}
\end{equation}
Based on the reparameterization trick, we can introduce an auxiliary denoising objective, which encourages the network to predict noiseless token $x_0$:
\begin{equation}
	\mathcal{L}_{x_0} = -\log p_\theta(\bm{x}_0|\bm{x}_t,\bm{y})
\label{eqn:L_x_0}
\end{equation}
We find that combining this loss with $\mathcal{L}_{vlb}$ improves the image quality.

\begin{algorithm}[t]
	\caption{Inference of the VQ-Diffusion, given fast inference time stride $\Delta_t$, input text $s$.} \label{alg:inference}
	\begin{algorithmic}[1]
		\State $t \gets T$, $\bm{y} \gets \text{BPE}(s)$
		\State $\bm{x}_t \gets \text{sample from } p(\bm{x}_T)$ \Comment{Eqn.~\ref{eqn:prior_p_x_T}}
		\While{$t > 0$}
		\State $\bm{x}_{t} \gets  \text{sample from}~p_\theta(\bm{x}_{t-\Delta_t}|\bm{x}_t, \bm{y})$ \Comment{Eqn.~\ref{eqn:fast_inference}}
		\State $t \gets t - \Delta_t$ 
		\EndWhile
		\State \Return VQVAE-Decoder($\bm{x}_t$)
	\end{algorithmic}
\end{algorithm}

\noindent\textbf{Model architecture.}
We propose an encoder-decoder transformer to estimate the distribution $p_\theta(\tilde{\bm{x}}_0 | \bm{x}_t,\bm{y})$. As shown in Figure~\ref{fig:method}, the framework contains two parts: a text encoder and a diffusion image decoder. Our text encoder takes the text tokens $\bm{y}$ and yields a conditional feature sequence. The diffusion image decoder takes the image token $\bm{x_t}$ and timestep $t$ and outputs the noiseless token distribution $p_\theta(\tilde{\bm{x}}_0 | \bm{x}_t,\bm{y})$. The decoder contains several transformer blocks and a softmax layer. Each transformer block contains a full attention, a cross attention to combine text information and a feed forward network block. The current timestep $t$ is injected into the network with Adaptive Layer Normalization~\cite{ba2016layer}(AdaLN) operator, \ie, $\text{AdaLN}(h, t) = a_t\text{LayerNorm}(h) + b_t$, where $h$ is the intermediate activations, $a_t$ and $b_t$ are obtained from a linear projection of the timestep embedding. 

\begin{figure*}[t!]
	\centering
	\vspace{-0.4cm}
	\includegraphics[width=1.0\linewidth]{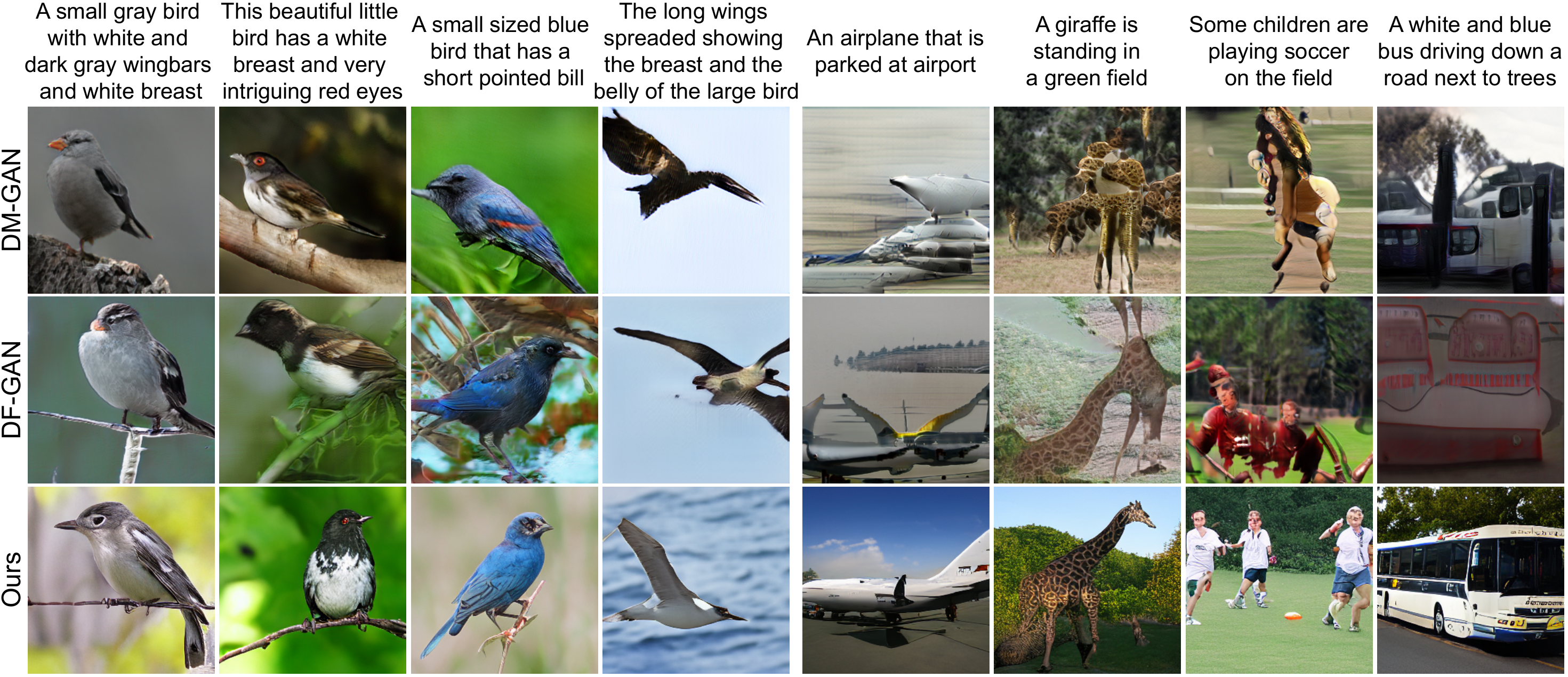}
	\vspace{-0.7cm}
	\caption{Comparison with GAN-based method on CUB-200 and MSCOCO datasets.}
	\label{fig:comparison}
	\vspace{-0.3cm}
\end{figure*}

\noindent\textbf{Fast inference strategy}
In the inference stage, by leveraging the reparameterization trick, we can skip some steps in diffusion model to achieve a faster inference. 

Specifically, assuming the time stride is $\Delta_t$, instead of sampling images in the chain of $x_T, x_{T-1},x_{T-2} ... x_0$, we sample images in the chain of  $x_T, x_{T-\Delta_t},x_{T-2\Delta_t} ... x_0$ with the reverse transition distribution:
\begin{equation}
	p_\theta(\bm{x}_{t-{\Delta_t}}|\bm{x}_t,\bm{y}) = \sum_{\tilde{\bm{x}}_0=1}^{K} q(\bm{x}_{t-{\Delta_t}}|\bm{x}_t,\tilde{\bm{x}}_0) p_\theta(\tilde{\bm{x}}_0|\bm{x}_t, \bm{y}).
	\label{eqn:fast_inference}
\end{equation}

We found it makes the sampling more efficient which only causes little harm to quality. The whole training and inference algorithm is shown in Algorithm~\ref{alg:training} and \ref{alg:inference}.

\section{Experiments}

In this section, we first introduce the overall experiment setups and then present extensive results to demonstrate the superiority of our approach in text-to-image synthesis. Finally, we point out that our method is a general image synthesis framework that achieves great performance on other generation tasks, including unconditional and class conditional image synthesis.

\noindent\textbf{Datasets.} To demonstrate the capability of our proposed method for text-to-image synthesis, we conduct experiments on CUB-200~\cite{wah2011caltech}, Oxford-102~\cite{nilsback2008automated}, and MSCOCO~\cite{lin2014microsoft} datasets. The CUB-200 dataset contains 8855 training images and 2933 test images belonging to 200 bird species. Oxford-102 dataset contains 8189 images of flowers of $102$ categories. Each image in CUB-200 and Oxford-102 dataset contains 10 text descriptions. MSCOCO dataset contains $82k$ images for training and $40k$ images for testing. Each image in this dataset has five text descriptions.

To further demonstrate the scalability of our method, we also train our model on large scale datasets, including Conceptual Captions~\cite{sharma2018conceptual,changpinyo2021conceptual} and LAION-400M~\cite{schuhmann2021laion}. The Conceptual Caption dataset, including both CC3M~\cite{sharma2018conceptual} and CC12M~\cite{changpinyo2021conceptual} datasets, contains 15M images. To balance the text and image distribution, we filter a 7M subset according to the word frequency. The LAION-400M dataset contains 400M image-text pairs. We train our model on three subsets from LAION, i.e., cartoon, icon, and human, each of them contains 0.9M, 1.3M, 42M images, respectively. For each subset, we filter the data according to the text.

\noindent\textbf{Traning Details.} Our VQ-VAE's encoder and decoder follow the setting of VQGAN~\cite{esser2021taming} which leverages the GAN loss to get a more realistic image. We directly adopt the publicly available VQGAN model trained on OpenImages~\cite{krasin2017openimages} dataset for all text-to-image synthesis experiments. It converts $256\times 256$ images into $32 \times 32$ tokens. The codebook size $K = 2886$ after removing useless codes. We adopt a publicly available tokenizer of the CLIP model~\cite{radford2021learning} as text encoder, yielding a conditional sequence of length 77. We fix both image and text encoders in our training. 

For fair comparison with previous text-to-image methods under similar parameters, we build two different diffusion image decoder settings: 1) \textbf{VQ-Diffusion-S} (Small), it contains $18$ transformer blocks with dimension of $192$. The model contains $34M$ parameters. 2) \textbf{VQ-Diffusion-B} (Base), it contains $19$ transformer blocks with dimension of $1024$. The model contains $370M$ parameters. 

In order to show the scalability of our method, we also train our base model on a larger database Conceptual Captions, and then fine-tune it on each database. This model is denoted as \textbf{VQ-Diffusion-F}.

For the default setting, we set timesteps $T = 100$ and loss weight $\lambda = 0.0005$. For the transition matrix, we linearly increase $\overline\gamma_t$ and $\overline\beta_t$ from $0$ to $0.9$ and $0.1$, respectively. We optimize our network using AdamW~\cite{loshchilov2017decoupled} with $\beta_1=0.9$ and $\beta_2=0.96$. The learning rate is set to $0.00045$ after 5000 iterations of warmup. More training details are provided in the appendix.

\subsection{Comparison with state-of-the-art methods}

We qualitatively compare the proposed method with several state-of-the-art methods, including some GAN-based methods~\cite{xu2018attngan,zhang2017stackgan,souza2020efficient,tan2019semantics,zhang2018stackgan++,zhu2019dm,tao2020df}, DALL-E~\cite{ramesh2021zero} and CogView~\cite{ding2021cogview}, on MSCOCO, CUB-200 and Oxford-102 datasets. We calculate the FID~\cite{heusel2017gans} between $30k$ generated images and $30k$ real images, and show the results in Table~\ref{table:text2image}.

We can see that our small model, VQ-Diffusion-S, which has the similar parameter number with previous GAN-based models, has strong performance on CUB-200 and Oxford-102 datasets. Our base model, VQ-Diffusion-B, further improves the performance. And our VQ-Diffusion-F model achieves the best results and surpasses all previous methods by a large margin, even surpassing DALL-E~\cite{ramesh2021zero} and CogView~\cite{ding2021cogview}, which have ten times more parameters than ours, on MSCOCO dataset.

Some visualized comparison results with DM-GAN~\cite{zhu2019dm} and DF-GAN~\cite{tao2020df} are shown in Figure~\ref{fig:comparison}. Obviously, our synthesized images have better realistic fine-grained details and are more consistent with the input text.

\subsection{In the wild text-to-image synthesis}
\begin{figure}[t!]
	\centering
	\vspace{-0.2cm}
	\includegraphics[width=1.0\linewidth]{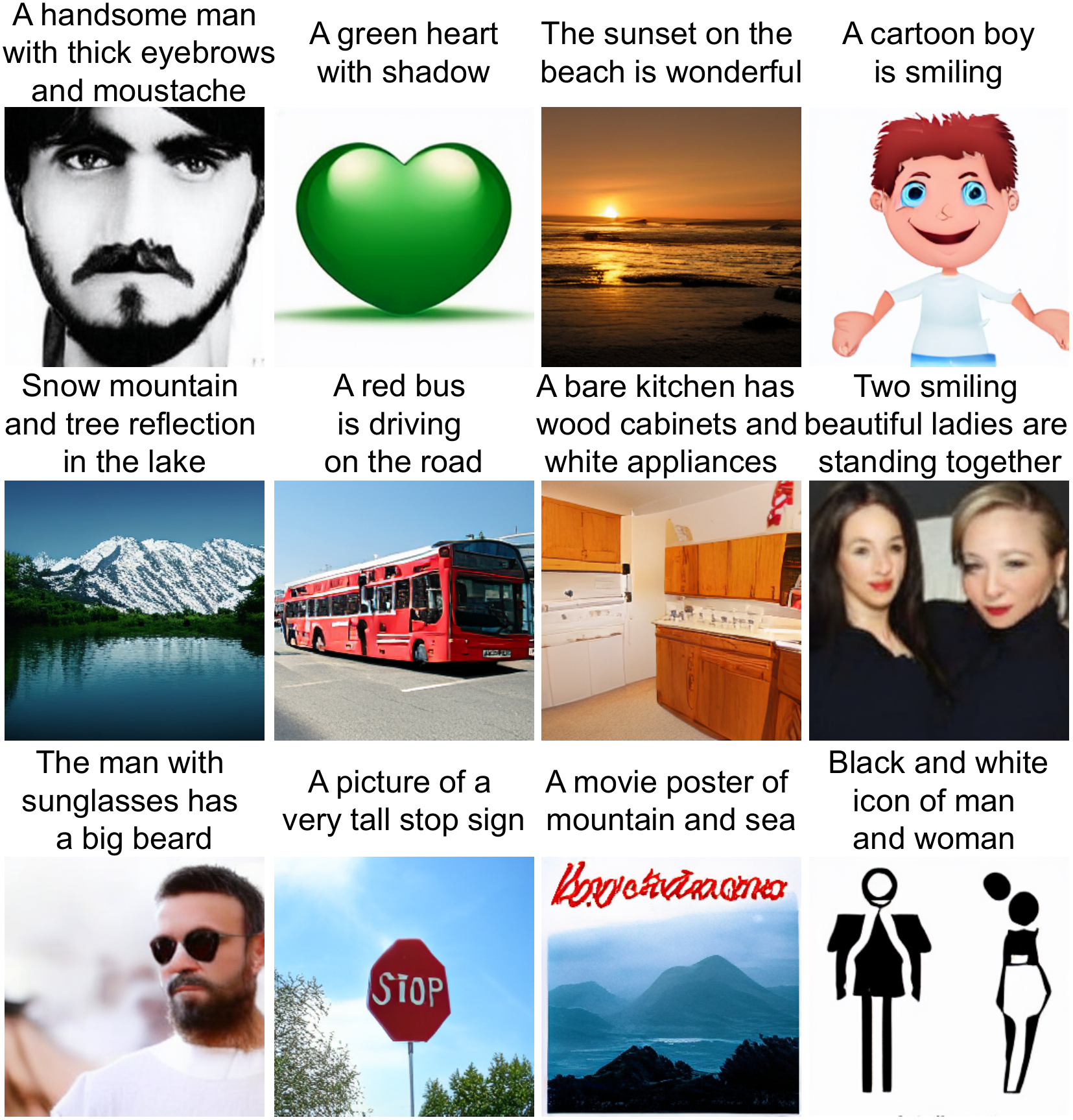}
	\vspace{-0.6cm}
	\caption{In the wild text-to-image synthesis results.}
	\vspace{-0.3cm}
	\label{fig:ourresult}
\end{figure}

To demonstrate the capability of generating in-the-wild images, we train our model on three subsets from LAION-400M dataset, \eg, cartoon, icon and human. We provide our results here in Figure~\ref{fig:ourresult}. Though our base model is much smaller than previous works like DALL-E and CogView, we also achieved a strong performance.

Compared with the AR method which generates images from top-left to down-right, our method generates images in a global manner. It makes our method can be applied to many vision tasks, \eg, irregular mask inpainting. For this task, we do not need to re-train a new model. We simply set the tokens in the irregular region as [MASK] token, and send them to our model.  
This strategy supports both unconditional mask inpainting and text conditional mask inpainting. We show these results in the appendix.

\begin{table}[t]
\vspace{-0.2cm}
\begin{tabular}{l|c|c|c}
    \hline
                  &  MSCOCO &  CUB-200  & Oxford-102  \\ 
	\shline
   \!\!\!\!  StackGAN~\cite{zhang2017stackgan}     &  74.05     & 51.89  &  55.28 \\
   \!\!\!\!  StackGAN++~\cite{zhang2018stackgan++}     &    81.59  &  15.30  & 48.68 \\
   \!\!\!\! EFF-T2I~\cite{souza2020efficient}       &    -    &     11.17     &  16.47 \\
   \!\!\!\!  SEGAN~\cite{tan2019semantics}     &     32.28     &      18.17     &  -  \\
   \!\!\!\!  AttnGAN~\cite{xu2018attngan}     & 35.49     &  23.98 &  - \\
   \!\!\!\!  DM-GAN~\cite{zhu2019dm}       & 32.64     &  16.09 &  - \\
   \!\!\!\!  DF-GAN~\cite{tao2020df}       & 21.42     &  14.81  &  -  \\
   \!\!\!\!  DAE-GAN~\cite{ruan2021dae}    &    28.12    &      15.19     &   -    \\
   \!\!\!\!  DALLE~\cite{ramesh2021zero}           & 27.50       &     56.10       &  -   \\
   \!\!\!\!  Cogview~\cite{ding2021cogview}       & 27.10       &       -     &   -   \\
\hline
   \!\!\!\!  VQ-Diffusion-S     &      30.17       &       12.97    &   14.95 \\
   \!\!\!\!  VQ-Diffusion-B     &    19.75     &     11.94      &  14.88  \\
   \!\!\!\!  VQ-Diffusion-F & \textbf{13.86}  &  \textbf{10.32} & \textbf{14.10} \\
	 \shline
\end{tabular}
\vspace{-0.3cm}
\caption{FID comparison of different text-to-image synthesis method on MSCOCO, CUB-200, and Oxford-102 datasets.}
\vspace{-0.2cm}
\label{table:text2image}
\end{table}

\subsection{Ablations}

\noindent \textbf{Number of timesteps.} We investigate the timesteps in training and inference. As shown in Table~\ref{table:fastinf}, we perform the experiment on the CUB-200 dataset. We find when the training steps increase from $10$ to $100$, the result improves, when it further increase to $200$, it seems saturated. So we set the default timesteps number to $100$ in our experiments. To demonstrate the fast inference strategy, we evaluate the generated images from $10,25,50,100$ inference steps on five models with different training steps. We find it still maintains a good performance when dropping $3/4$ inference steps, which may save about $3/4$ inference times.

\begin{figure}[t!]
	\centering
	\vspace{-0.2cm}
	\includegraphics[width=1.0\linewidth]{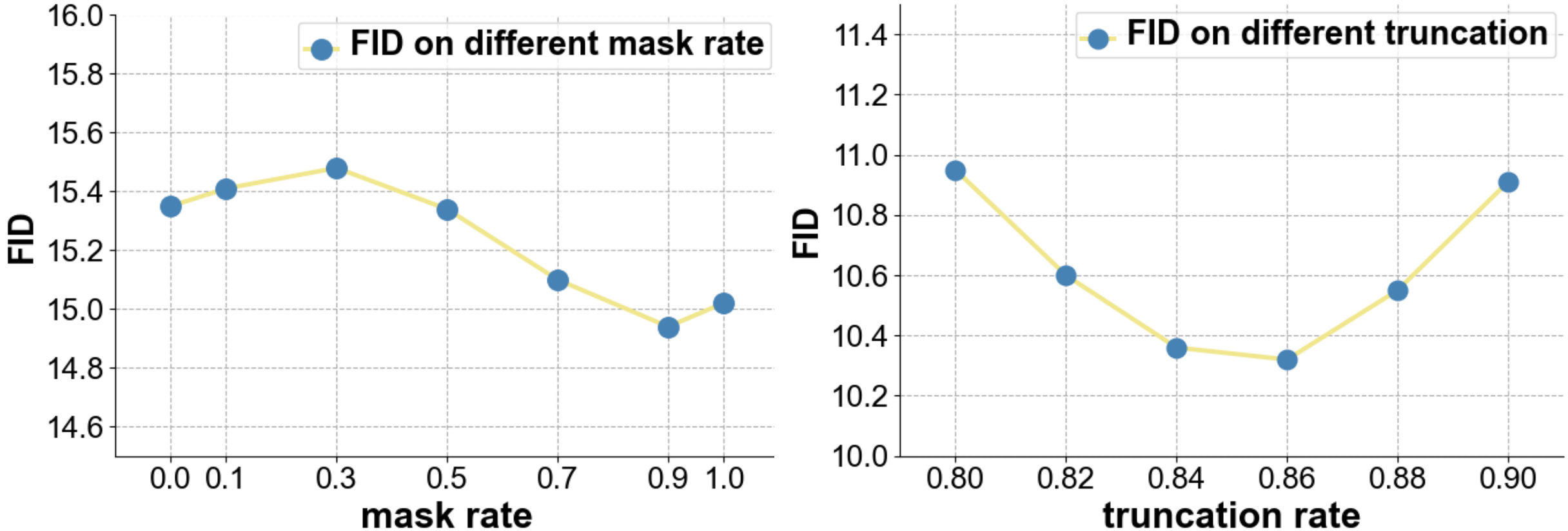}
	\vspace{-0.6cm}
	\caption{Ablation study on the mask rate and the truncation rate.}
	\label{fig:ablation}
\end{figure}

\noindent\textbf{Mask-and-replace diffusion strategy.} We explore how the mask-and-replace strategy benefits our performance on the Oxford-102 dataset. We set different final mask rate ($\overline\gamma_T$) to investigate the effect. Both mask only strategy ($\overline\gamma_T=1$) and replace only strategy ($\overline\gamma_T=0$) are special cases of our mask-and-replace strategy. From Figure~\ref{fig:ablation}, we find it get the best performance when $M=0.9$. When $M>0.9$, it may suffer from the error accumulation problem, when $M<0.9$, the network may be difficult to find which region needs to pay more attention.

\begin{table}[t]
\centering
\vspace{-0.1cm}
\begin{tabular}{c|c|c|c|c|c|c}
\hline
\multirow{7}{*}{\rotatebox{90}{\!\!\!\!\!\!\!\!\!\!\!\!\!\! inference steps}}&
\multicolumn{6}{c}{training steps} \\
\shline
&  & 10&25&50&100&200\\
\cline{2-7}
  & 10 &32.35 &27.62  & 23.47 & 19.84  &  20.96 \\
\cline{2-7}
  & 25 & -        &18.53&  15.25 & 14.03  &  16.13 \\
\cline{2-7}
  & 	 50 & -        & -      &13.82 & 12.45 & 13.67   \\
\cline{2-7}
  & 	 100 & -      & -      & -        & 11.94 & 12.27 \\
\cline{2-7}
  & 	 200 & -      & -      & -        & -         & 11.80 \\
\shline
\end{tabular}
\vspace{-0.3cm}
\caption{Ablation study on training steps and inference steps. Each column shares the same training steps while each row shares the same inference steps.}
\label{table:fastinf}
\end{table}

\begin{table}[t]
\centering
\begin{tabular}{l|c|c|c}
    \hline
    Model  & steps &  FID & throughput(imgs/s)   \\ 
	\shline
	 VQ-AR-S & & 18.12 & 0.08 \\
	 VQ-Diffusion-S & 25 &  15.46 & 1.25 \\
	 VQ-Diffusion-S & 50 & 13.62  & 0.67 \\
	 VQ-Diffusion-S & 100 &  12.97  & 0.37 \\
          \shline
	 VQ-AR-B & & 17.76   & 0.03 \\
	 VQ-Diffusion-B & 25 & 14.03  & 0.47 \\
	 VQ-Diffusion-B & 50 & 12.45  & 0.24 \\
	 VQ-Diffusion-B & 100 & 11.94  & 0.13 \\
          \shline
\end{tabular}
\vspace{-0.2cm}
\caption{Comparison between VQ-Diffusion and VQ-AR models. By changing the inference steps, the VQ-Diffusion model is $15$ times faster than the VQ-AR model while maintaining better performance.}
\label{table:autoregressive}
\vspace{-0.3cm}
\end{table}

\noindent \textbf{Truncation.} We also demonstrate that the truncation sampling strategy is extremely important for our discrete diffusion based method. It may avoid the network sampling from low probability tokens. Specifically, we only keep top $r$ tokens of $p_\theta(\tilde{\bm{x}}_0|\bm{x}_t, \bm{y})$ in the inference stage. We evaluate the results with different truncation rates $r$ on CUB-200 dataset. As shown in Figure~\ref{fig:ablation}, we find that it achieves the best performance when the truncation rate equals $0.86$.

\noindent \textbf{VQ-Diffusion vs VQ-AR.} For a fair comparison, we replace our diffusion image decoder with an autoregressive decoder with the same network structure and keep other settings the same, including both image and text encoders. The autoregressive model is denoted as VQ-AR-S and VQ-AR-B, corresponding to VQ-Diffusion-S and VQ-Diffusion-B. The experiment is performed on the CUB-200 dataset. As shown in Table~\ref{table:autoregressive} , on both -S and -B settings the VQ-Diffusion model surpasses the VQ-AR model by a large margin. Meanwhile, we evaluate the throughput of both methods on a V100 GPU with a batch size of 32. The VQ-Diffusion with the fast inference strategy is $15$ times faster than the VQ-AR model with a better FID score.

\subsection{Unified generation model}
Our method is general, which can also be applied to other image synthesis tasks, \eg, unconditional image synthesis and image synthesis conditioned on labels. To generate images from a given class label, we first remove the text encoder network and cross attention part in transformer blocks, and inject the class label through the AdaLN operator. Our network contains $24$ transformer blocks with dimension $512$. We train our model on the ImageNet dataset. For VQ-VAE, we adopt the publicly available model from VQ-GAN~\cite{esser2021taming} trained on ImageNet dataset, which downsamples images from $256 \times 256$ to $16 \times 16$. For unconditional image synthesis, we trained our model on the FFHQ256 dataset, which contains 70k high quality face images. The image encoder also downsamples images to $16 \times 16$ tokens. 

We assess the performance of our model in terms of FID and compare with a variety of previously established models~\cite{brock2018large,nichol2021improved,dhariwal2021diffusion,esser2021taming,esser2021imagebart}. For a fair comparison, we calculate FID between $50k$ generated images and all real images. Following~\cite{esser2021taming} we can further increase the quality by only accepting images with a top $5\%$ classification score, denoted as acc0.05. We show the quantitative results in Table ~\ref{table:imagenet}. While some task-specialized GAN models report better FID scores, our approach provides a unified model that works well across a wide range of tasks.

\begin{table}[t]
\centering
\vspace{-0.2cm}
\begin{tabular}{l|c|c}
    \hline
             Model  &  ImageNet & FFHQ  \\ 
	\shline
	 StyleGAN2~\cite{karras2020analyzing} & - & 3.8 \\
	 BigGAN~\cite{brock2018large} & 7.53 & 12.4 \\
	 BigGAN-deep~\cite{brock2018large} &  6.84 & - \\
	 IDDPM~\cite{nichol2021improved} & 12.3 & - \\
	 ADM-G~\cite{dhariwal2021diffusion} &  10.94 & - \\
     VQGAN~\cite{esser2021taming} &  15.78 & 9.6 \\
     ImageBART~\cite{esser2021imagebart} & 21.19 & 9.57 \\
	 Ours &  11.89 & 6.33 \\
\shline
	 ADM-G (1.0guid)~\cite{dhariwal2021diffusion} &  4.59 & - \\
	 VQGAN (acc0.05)~\cite{esser2021taming} &  5.88 &  - \\
     ImageBART (acc0.05)~\cite{esser2021imagebart} & 7.44 & - \\ 
	 Ours (acc0.05) & 5.32  & - \\
	 \shline
\end{tabular}
\vspace{-0.2cm}
\caption{FID score comparison for class-conditional synthesis on ImageNet, and unconditional synthesis on FFHQ dataset. 'guid' denotes using classifier guidance~\cite{dhariwal2021diffusion}, 'acc' denotes adopting acceptance rate~\cite{esser2021taming}.}
\label{table:imagenet}
\vspace{-0.4cm}
\end{table}

\section{Conclusion}

In this paper, we present a novel text-to-image architecture named VQ-Diffusion. The core design is to model the VQ-VAE latent space in a non-autoregressive manner. The proposed mask-and-replace diffusion strategy avoids the accumulation of errors of the AR model. Our model has the capacity to generate more complex scenes, which surpasses previous GAN-based text-to-image methods. Our method is also general and produces strong results on unconditional and conditional image generation.

\section*{Acknowledgement}
We thank Qiankun Liu from University of Science and Technology of China for his help, he provided the initial code and datasets.

{\small
\bibliographystyle{ieee_fullname}
\bibliography{egbib}

\begin{thebibliography}{10}\itemsep=-1pt

\bibitem{austin2021structured}
Jacob Austin, Daniel~D Johnson, Jonathan Ho, Daniel Tarlow, and Rianne van~den
  Berg.
\newblock Structured denoising diffusion models in discrete state-spaces.
\newblock {\em arXiv preprint arXiv:2107.03006}, 2021.

\bibitem{ba2016layer}
Jimmy~Lei Ba, Jamie~Ryan Kiros, and Geoffrey~E Hinton.
\newblock Layer normalization.
\newblock {\em arXiv preprint arXiv:1607.06450}, 2016.

\bibitem{brock2018large}
Andrew Brock, Jeff Donahue, and Karen Simonyan.
\newblock Large scale gan training for high fidelity natural image synthesis.
\newblock {\em arXiv preprint arXiv:1809.11096}, 2018.

\bibitem{brown2020language}
Tom~B Brown, Benjamin Mann, Nick Ryder, Melanie Subbiah, Jared Kaplan, Prafulla
  Dhariwal, Arvind Neelakantan, Pranav Shyam, Girish Sastry, Amanda Askell,
  et~al.
\newblock Language models are few-shot learners.
\newblock {\em arXiv preprint arXiv:2005.14165}, 2020.

\bibitem{cha2019adversarial}
Miriam Cha, Youngjune~L Gwon, and HT Kung.
\newblock Adversarial learning of semantic relevance in text to image
  synthesis.
\newblock In {\em Proceedings of the AAAI conference on artificial
  intelligence}, volume~33, pages 3272--3279, 2019.

\bibitem{changpinyo2021conceptual}
Soravit Changpinyo, Piyush Sharma, Nan Ding, and Radu Soricut.
\newblock Conceptual 12m: Pushing web-scale image-text pre-training to
  recognize long-tail visual concepts.
\newblock In {\em Proceedings of the IEEE/CVF Conference on Computer Vision and
  Pattern Recognition}, pages 3558--3568, 2021.

\bibitem{chen2020generative}
Mark Chen, Alec Radford, Rewon Child, Jeffrey Wu, Heewoo Jun, David Luan, and
  Ilya Sutskever.
\newblock Generative pretraining from pixels.
\newblock In {\em International Conference on Machine Learning}, pages
  1691--1703. PMLR, 2020.

\bibitem{cheng2020rifegan}
Jun Cheng, Fuxiang Wu, Yanling Tian, Lei Wang, and Dapeng Tao.
\newblock Rifegan: Rich feature generation for text-to-image synthesis from
  prior knowledge.
\newblock In {\em Proceedings of the IEEE/CVF Conference on Computer Vision and
  Pattern Recognition}, pages 10911--10920, 2020.

\bibitem{dash2017tac}
Ayushman Dash, John Cristian~Borges Gamboa, Sheraz Ahmed, Marcus Liwicki, and
  Muhammad~Zeshan Afzal.
\newblock Tac-gan-text conditioned auxiliary classifier generative adversarial
  network.
\newblock {\em arXiv preprint arXiv:1703.06412}, 2017.

\bibitem{deng2009imagenet}
Jia Deng, Wei Dong, Richard Socher, Li-Jia Li, Kai Li, and Li Fei-Fei.
\newblock Imagenet: A large-scale hierarchical image database.
\newblock In {\em 2009 IEEE conference on computer vision and pattern
  recognition}, pages 248--255. Ieee, 2009.

\bibitem{devlin2018bert}
Jacob Devlin, Ming-Wei Chang, Kenton Lee, and Kristina Toutanova.
\newblock Bert: Pre-training of deep bidirectional transformers for language
  understanding.
\newblock {\em arXiv preprint arXiv:1810.04805}, 2018.

\bibitem{dhariwal2021diffusion}
Prafulla Dhariwal and Alex Nichol.
\newblock Diffusion models beat gans on image synthesis.
\newblock {\em arXiv preprint arXiv:2105.05233}, 2021.

\bibitem{ding2021cogview}
Ming Ding, Zhuoyi Yang, Wenyi Hong, Wendi Zheng, Chang Zhou, Da Yin, Junyang
  Lin, Xu Zou, Zhou Shao, Hongxia Yang, et~al.
\newblock Cogview: Mastering text-to-image generation via transformers.
\newblock {\em arXiv preprint arXiv:2105.13290}, 2021.

\bibitem{el2019tell}
Alaaeldin El-Nouby, Shikhar Sharma, Hannes Schulz, Devon Hjelm, Layla~El Asri,
  Samira~Ebrahimi Kahou, Yoshua Bengio, and Graham~W Taylor.
\newblock Tell, draw, and repeat: Generating and modifying images based on
  continual linguistic instruction.
\newblock In {\em Proceedings of the IEEE/CVF International Conference on
  Computer Vision}, pages 10304--10312, 2019.

\bibitem{esser2021imagebart}
Patrick Esser, Robin Rombach, Andreas Blattmann, and Bj{\"o}rn Ommer.
\newblock Imagebart: Bidirectional context with multinomial diffusion for
  autoregressive image synthesis.
\newblock {\em arXiv preprint arXiv:2108.08827}, 2021.

\bibitem{esser2021taming}
Patrick Esser, Robin Rombach, and Bjorn Ommer.
\newblock Taming transformers for high-resolution image synthesis.
\newblock In {\em Proceedings of the IEEE/CVF Conference on Computer Vision and
  Pattern Recognition}, pages 12873--12883, 2021.

\bibitem{gao2019perceptual}
Lianli Gao, Daiyuan Chen, Jingkuan Song, Xing Xu, Dongxiang Zhang, and Heng~Tao
  Shen.
\newblock Perceptual pyramid adversarial networks for text-to-image synthesis.
\newblock In {\em Proceedings of the AAAI Conference on Artificial
  Intelligence}, volume~33, pages 8312--8319, 2019.

\bibitem{goodfellow2014generative}
Ian Goodfellow, Jean Pouget-Abadie, Mehdi Mirza, Bing Xu, David Warde-Farley,
  Sherjil Ozair, Aaron Courville, and Yoshua Bengio.
\newblock Generative adversarial nets.
\newblock In {\em Advances in Neural Information Processing Systems}, pages
  2672--2680, 2014.

\bibitem{gu2020giqa}
Shuyang Gu, Jianmin Bao, Dong Chen, and Fang Wen.
\newblock Giqa: Generated image quality assessment.
\newblock In {\em European Conference on Computer Vision}, pages 369--385.
  Springer, 2020.

\bibitem{gu2020priorgan}
Shuyang Gu, Jianmin Bao, Dong Chen, and Fang Wen.
\newblock Priorgan: Real data prior for generative adversarial nets.
\newblock {\em arXiv preprint arXiv:2006.16990}, 2020.

\bibitem{gu2019mask}
Shuyang Gu, Jianmin Bao, Hao Yang, Dong Chen, Fang Wen, and Lu Yuan.
\newblock Mask-guided portrait editing with conditional gans.
\newblock In {\em Proceedings of the IEEE/CVF Conference on Computer Vision and
  Pattern Recognition}, pages 3436--3445, 2019.

\bibitem{heusel2017gans}
Martin Heusel, Hubert Ramsauer, Thomas Unterthiner, Bernhard Nessler, and Sepp
  Hochreiter.
\newblock Gans trained by a two time-scale update rule converge to a local nash
  equilibrium.
\newblock {\em Advances in neural information processing systems}, 30, 2017.

\bibitem{ho2020denoising}
Jonathan Ho, Ajay Jain, and Pieter Abbeel.
\newblock Denoising diffusion probabilistic models.
\newblock {\em arXiv preprint arXiv:2006.11239}, 2020.

\bibitem{ho2021cascaded}
Jonathan Ho, Chitwan Saharia, William Chan, David~J Fleet, Mohammad Norouzi,
  and Tim Salimans.
\newblock Cascaded diffusion models for high fidelity image generation.
\newblock {\em arXiv preprint arXiv:2106.15282}, 2021.

\bibitem{hong2018inferring}
Seunghoon Hong, Dingdong Yang, Jongwook Choi, and Honglak Lee.
\newblock Inferring semantic layout for hierarchical text-to-image synthesis.
\newblock In {\em Proceedings of the IEEE Conference on Computer Vision and
  Pattern Recognition}, pages 7986--7994, 2018.

\bibitem{hoogeboom2021argmax}
Emiel Hoogeboom, Didrik Nielsen, Priyank Jaini, Patrick Forr{\'e}, and Max
  Welling.
\newblock Argmax flows and multinomial diffusion: Towards non-autoregressive
  language models.
\newblock {\em arXiv preprint arXiv:2102.05379}, 2021.

\bibitem{huang2021unifying}
Yupan Huang, Hongwei Xue, Bei Liu, and Yutong Lu.
\newblock Unifying multimodal transformer for bi-directional image and text
  generation.
\newblock In {\em Proceedings of the 29th ACM International Conference on
  Multimedia}, pages 1138--1147, 2021.

\bibitem{karras2019style}
Tero Karras, Samuli Laine, and Timo Aila.
\newblock A style-based generator architecture for generative adversarial
  networks.
\newblock In {\em Proceedings of the IEEE/CVF Conference on Computer Vision and
  Pattern Recognition}, pages 4401--4410, 2019.

\bibitem{karras2020analyzing}
Tero Karras, Samuli Laine, Miika Aittala, Janne Hellsten, Jaakko Lehtinen, and
  Timo Aila.
\newblock Analyzing and improving the image quality of stylegan.
\newblock In {\em Proceedings of the IEEE/CVF Conference on Computer Vision and
  Pattern Recognition}, pages 8110--8119, 2020.

\bibitem{krasin2017openimages}
Ivan Krasin, Tom Duerig, Neil Alldrin, Vittorio Ferrari, Sami Abu-El-Haija,
  Alina Kuznetsova, Hassan Rom, Jasper Uijlings, Stefan Popov, Andreas Veit,
  et~al.
\newblock Openimages: A public dataset for large-scale multi-label and
  multi-class image classification.
\newblock {\em Dataset available from https://github. com/openimages}, 2(3):18,
  2017.

\bibitem{lao2019dual}
Qicheng Lao, Mohammad Havaei, Ahmad Pesaranghader, Francis Dutil, Lisa~Di
  Jorio, and Thomas Fevens.
\newblock Dual adversarial inference for text-to-image synthesis.
\newblock In {\em Proceedings of the IEEE/CVF International Conference on
  Computer Vision}, pages 7567--7576, 2019.

\bibitem{li2019controllable}
Bowen Li, Xiaojuan Qi, Thomas Lukasiewicz, and Philip~HS Torr.
\newblock Controllable text-to-image generation.
\newblock {\em arXiv preprint arXiv:1909.07083}, 2019.

\bibitem{li2019object}
Wenbo Li, Pengchuan Zhang, Lei Zhang, Qiuyuan Huang, Xiaodong He, Siwei Lyu,
  and Jianfeng Gao.
\newblock Object-driven text-to-image synthesis via adversarial training.
\newblock In {\em Proceedings of the IEEE/CVF Conference on Computer Vision and
  Pattern Recognition}, pages 12174--12182, 2019.

\bibitem{liang2020cpgan}
Jiadong Liang, Wenjie Pei, and Feng Lu.
\newblock Cpgan: Content-parsing generative adversarial networks for
  text-to-image synthesis.
\newblock In {\em European Conference on Computer Vision}, pages 491--508.
  Springer, 2020.

\bibitem{lin2021m6}
Junyang Lin, Rui Men, An Yang, Chang Zhou, Ming Ding, Yichang Zhang, Peng Wang,
  Ang Wang, Le Jiang, Xianyan Jia, et~al.
\newblock M6: A chinese multimodal pretrainer.
\newblock {\em arXiv preprint arXiv:2103.00823}, 2021.

\bibitem{lin2014microsoft}
Tsung-Yi Lin, Michael Maire, Serge Belongie, James Hays, Pietro Perona, Deva
  Ramanan, Piotr Doll{\'a}r, and C~Lawrence Zitnick.
\newblock Microsoft coco: Common objects in context.
\newblock In {\em European conference on computer vision}, pages 740--755.
  Springer, 2014.

\bibitem{loshchilov2017decoupled}
Ilya Loshchilov and Frank Hutter.
\newblock Decoupled weight decay regularization.
\newblock {\em arXiv preprint arXiv:1711.05101}, 2017.

\bibitem{nguyen2017plug}
Anh Nguyen, Jeff Clune, Yoshua Bengio, Alexey Dosovitskiy, and Jason Yosinski.
\newblock Plug \& play generative networks: Conditional iterative generation of
  images in latent space.
\newblock In {\em Proceedings of the IEEE Conference on Computer Vision and
  Pattern Recognition}, pages 4467--4477, 2017.

\bibitem{nichol2021improved}
Alex Nichol and Prafulla Dhariwal.
\newblock Improved denoising diffusion probabilistic models.
\newblock {\em arXiv preprint arXiv:2102.09672}, 2021.

\bibitem{nilsback2008automated}
Maria-Elena Nilsback and Andrew Zisserman.
\newblock Automated flower classification over a large number of classes.
\newblock In {\em 2008 Sixth Indian Conference on Computer Vision, Graphics \&
  Image Processing}, pages 722--729. IEEE, 2008.

\bibitem{oord2017neural}
Aaron van~den Oord, Oriol Vinyals, and Koray Kavukcuoglu.
\newblock Neural discrete representation learning.
\newblock {\em arXiv preprint arXiv:1711.00937}, 2017.

\bibitem{parmar2018image}
Niki Parmar, Ashish Vaswani, Jakob Uszkoreit, Lukasz Kaiser, Noam Shazeer,
  Alexander Ku, and Dustin Tran.
\newblock Image transformer.
\newblock In {\em International Conference on Machine Learning}, pages
  4055--4064. PMLR, 2018.

\bibitem{qiao2019learn}
Tingting Qiao, Jing Zhang, Duanqing Xu, and Dacheng Tao.
\newblock Learn, imagine and create: Text-to-image generation from prior
  knowledge.
\newblock {\em Advances in Neural Information Processing Systems}, 32:887--897,
  2019.

\bibitem{qiao2019mirrorgan}
Tingting Qiao, Jing Zhang, Duanqing Xu, and Dacheng Tao.
\newblock Mirrorgan: Learning text-to-image generation by redescription.
\newblock In {\em Proceedings of the IEEE/CVF Conference on Computer Vision and
  Pattern Recognition}, pages 1505--1514, 2019.

\bibitem{radford2021learning}
Alec Radford, Jong~Wook Kim, Chris Hallacy, Aditya Ramesh, Gabriel Goh,
  Sandhini Agarwal, Girish Sastry, Amanda Askell, Pamela Mishkin, Jack Clark,
  et~al.
\newblock Learning transferable visual models from natural language
  supervision.
\newblock {\em arXiv preprint arXiv:2103.00020}, 2021.

\bibitem{radford2018improving}
Alec Radford, Karthik Narasimhan, Tim Salimans, and Ilya Sutskever.
\newblock Improving language understanding by generative pre-training.
\newblock 2018.

\bibitem{radford2019language}
Alec Radford, Jeffrey Wu, Rewon Child, David Luan, Dario Amodei, Ilya
  Sutskever, et~al.
\newblock Language models are unsupervised multitask learners.
\newblock {\em OpenAI blog}, 1(8):9, 2019.

\bibitem{ramesh2021zero}
Aditya Ramesh, Mikhail Pavlov, Gabriel Goh, Scott Gray, Chelsea Voss, Alec
  Radford, Mark Chen, and Ilya Sutskever.
\newblock Zero-shot text-to-image generation.
\newblock {\em arXiv preprint arXiv:2102.12092}, 2021.

\bibitem{razavi2019generating}
Ali Razavi, Aaron van~den Oord, and Oriol Vinyals.
\newblock Generating diverse high-fidelity images with vq-vae-2.
\newblock In {\em Advances in neural information processing systems}, pages
  14866--14876, 2019.

\bibitem{reed2016generative}
Scott Reed, Zeynep Akata, Xinchen Yan, Lajanugen Logeswaran, Bernt Schiele, and
  Honglak Lee.
\newblock Generative adversarial text to image synthesis.
\newblock In {\em International Conference on Machine Learning}, pages
  1060--1069. PMLR, 2016.

\bibitem{ruan2021dae}
Shulan Ruan, Yong Zhang, Kun Zhang, Yanbo Fan, Fan Tang, Qi Liu, and Enhong
  Chen.
\newblock Dae-gan: Dynamic aspect-aware gan for text-to-image synthesis.
\newblock In {\em Proceedings of the IEEE/CVF International Conference on
  Computer Vision}, pages 13960--13969, 2021.

\bibitem{saharia2021image}
Chitwan Saharia, Jonathan Ho, William Chan, Tim Salimans, David~J Fleet, and
  Mohammad Norouzi.
\newblock Image super-resolution via iterative refinement.
\newblock {\em arXiv preprint arXiv:2104.07636}, 2021.

\bibitem{salimans2017pixelcnn++}
Tim Salimans, Andrej Karpathy, Xi Chen, and Diederik~P Kingma.
\newblock Pixelcnn++: Improving the pixelcnn with discretized logistic mixture
  likelihood and other modifications.
\newblock {\em arXiv preprint arXiv:1701.05517}, 2017.

\bibitem{schmidt2019generalization}
Florian Schmidt.
\newblock Generalization in generation: A closer look at exposure bias.
\newblock {\em arXiv preprint arXiv:1910.00292}, 2019.

\bibitem{schuhmann2021laion}
Christoph Schuhmann, Richard Vencu, Romain Beaumont, Robert Kaczmarczyk,
  Clayton Mullis, Aarush Katta, Theo Coombes, Jenia Jitsev, and Aran
  Komatsuzaki.
\newblock Laion-400m: Open dataset of clip-filtered 400 million image-text
  pairs.
\newblock {\em arXiv preprint arXiv:2111.02114}, 2021.

\bibitem{sennrich2015neural}
Rico Sennrich, Barry Haddow, and Alexandra Birch.
\newblock Neural machine translation of rare words with subword units.
\newblock {\em arXiv preprint arXiv:1508.07909}, 2015.

\bibitem{sharma2018conceptual}
Piyush Sharma, Nan Ding, Sebastian Goodman, and Radu Soricut.
\newblock Conceptual captions: A cleaned, hypernymed, image alt-text dataset
  for automatic image captioning.
\newblock In {\em Proceedings of the 56th Annual Meeting of the Association for
  Computational Linguistics (Volume 1: Long Papers)}, pages 2556--2565, 2018.

\bibitem{sharma2018chatpainter}
Shikhar Sharma, Dendi Suhubdy, Vincent Michalski, Samira~Ebrahimi Kahou, and
  Yoshua Bengio.
\newblock Chatpainter: Improving text to image generation using dialogue.
\newblock {\em arXiv preprint arXiv:1802.08216}, 2018.

\bibitem{sohl2015deep}
Jascha Sohl-Dickstein, Eric Weiss, Niru Maheswaranathan, and Surya Ganguli.
\newblock Deep unsupervised learning using nonequilibrium thermodynamics.
\newblock In {\em International Conference on Machine Learning}, pages
  2256--2265. PMLR, 2015.

\bibitem{souza2020efficient}
Douglas~M Souza, J{\^o}natas Wehrmann, and Duncan~D Ruiz.
\newblock Efficient neural architecture for text-to-image synthesis.
\newblock In {\em 2020 International Joint Conference on Neural Networks
  (IJCNN)}, pages 1--8. IEEE, 2020.

\bibitem{tan2019semantics}
Hongchen Tan, Xiuping Liu, Xin Li, Yi Zhang, and Baocai Yin.
\newblock Semantics-enhanced adversarial nets for text-to-image synthesis.
\newblock In {\em Proceedings of the IEEE/CVF International Conference on
  Computer Vision}, pages 10501--10510, 2019.

\bibitem{tan2020kt}
Hongchen Tan, Xiuping Liu, Meng Liu, Baocai Yin, and Xin Li.
\newblock Kt-gan: knowledge-transfer generative adversarial network for
  text-to-image synthesis.
\newblock {\em IEEE Transactions on Image Processing}, 30:1275--1290, 2020.

\bibitem{tao2020df}
Ming Tao, Hao Tang, Songsong Wu, Nicu Sebe, Xiao-Yuan Jing, Fei Wu, and Bingkun
  Bao.
\newblock Df-gan: Deep fusion generative adversarial networks for text-to-image
  synthesis.
\newblock {\em arXiv preprint arXiv:2008.05865}, 2020.

\bibitem{van2016pixel}
Aaron Van~Oord, Nal Kalchbrenner, and Koray Kavukcuoglu.
\newblock Pixel recurrent neural networks.
\newblock In {\em International Conference on Machine Learning}, pages
  1747--1756. PMLR, 2016.

\bibitem{vaswani2017attention}
Ashish Vaswani, Noam Shazeer, Niki Parmar, Jakob Uszkoreit, Llion Jones,
  Aidan~N Gomez, {\L}ukasz Kaiser, and Illia Polosukhin.
\newblock Attention is all you need.
\newblock In {\em Advances in neural information processing systems}, pages
  5998--6008, 2017.

\bibitem{wah2011caltech}
Catherine Wah, Steve Branson, Peter Welinder, Pietro Perona, and Serge
  Belongie.
\newblock The caltech-ucsd birds-200-2011 dataset.
\newblock 2011.

\bibitem{xu2018attngan}
Tao Xu, Pengchuan Zhang, Qiuyuan Huang, Han Zhang, Zhe Gan, Xiaolei Huang, and
  Xiaodong He.
\newblock Attngan: Fine-grained text to image generation with attentional
  generative adversarial networks.
\newblock In {\em Proceedings of the IEEE conference on computer vision and
  pattern recognition}, pages 1316--1324, 2018.

\bibitem{yin2019semantics}
Guojun Yin, Bin Liu, Lu Sheng, Nenghai Yu, Xiaogang Wang, and Jing Shao.
\newblock Semantics disentangling for text-to-image generation.
\newblock In {\em Proceedings of the IEEE/CVF Conference on Computer Vision and
  Pattern Recognition}, pages 2327--2336, 2019.

\bibitem{zhang2021cross}
Han Zhang, Jing~Yu Koh, Jason Baldridge, Honglak Lee, and Yinfei Yang.
\newblock Cross-modal contrastive learning for text-to-image generation.
\newblock In {\em Proceedings of the IEEE/CVF Conference on Computer Vision and
  Pattern Recognition}, pages 833--842, 2021.

\bibitem{zhang2017stackgan}
Han Zhang, Tao Xu, Hongsheng Li, Shaoting Zhang, Xiaogang Wang, Xiaolei Huang,
  and Dimitris~N Metaxas.
\newblock Stackgan: Text to photo-realistic image synthesis with stacked
  generative adversarial networks.
\newblock In {\em Proceedings of the IEEE international conference on computer
  vision}, pages 5907--5915, 2017.

\bibitem{zhang2018stackgan++}
Han Zhang, Tao Xu, Hongsheng Li, Shaoting Zhang, Xiaogang Wang, Xiaolei Huang,
  and Dimitris~N Metaxas.
\newblock Stackgan++: Realistic image synthesis with stacked generative
  adversarial networks.
\newblock {\em IEEE transactions on pattern analysis and machine intelligence},
  41(8):1947--1962, 2018.

\bibitem{zhang2018photographic}
Zizhao Zhang, Yuanpu Xie, and Lin Yang.
\newblock Photographic text-to-image synthesis with a hierarchically-nested
  adversarial network.
\newblock In {\em Proceedings of the IEEE Conference on Computer Vision and
  Pattern Recognition}, pages 6199--6208, 2018.

\bibitem{zhu2019dm}
Minfeng Zhu, Pingbo Pan, Wei Chen, and Yi Yang.
\newblock Dm-gan: Dynamic memory generative adversarial networks for
  text-to-image synthesis.
\newblock In {\em Proceedings of the IEEE/CVF Conference on Computer Vision and
  Pattern Recognition}, pages 5802--5810, 2019.

\end{thebibliography}
}

\clearpage
\appendix

\section{Implementation details}
In our experiments on text-to-image synthesis, we adopt the public VQ-VAE~\cite{oord2017neural} model provided by VQGAN~\cite{esser2021taming} trained on the OpenImages~\cite{krasin2017openimages} dataset, which downsamples images from $256 \times 256$ to $32 \times 32$. We use the CLIP~\cite{radford2021learning} pretrained model (ViT-B) as our text encoder, which encodes a sentence to $77$ tokens. Our diffusion image decoder consists of several transformer blocks, each block contains full attention, cross attention, and feed forward network(FFN). Our base model contains $19$ transformer blocks, the channel of each block is $1024$. The FFN contains two linear layer, which expand the dimension to $4096$ in the middle layer. The model contains $370$M parameters. For our small model, it contains $18$ transformer blocks while the channel is $192$, the FFN contains two convolution layers with kernel size $3$, the channel expand rate is $2$. The model contains $34$M parameters.

For our class conditional generation model on ImageNet, we adopt the public VQ-VAE model provided by VQGAN trained on ImageNet, which downsamples images from $256 \times 256$ to $16 \times 16$. Our model contains $24$ transformer blocks, each block contains a full attention layer and a FFN. The base channel number is $512$. Besides, the FFN also uses convolution instead of linear layer, and the channel expand rate is $4$.

\begin{figure}[b]
	\centering
	\vspace{-0.4cm}
	\includegraphics[width=1.0\linewidth]{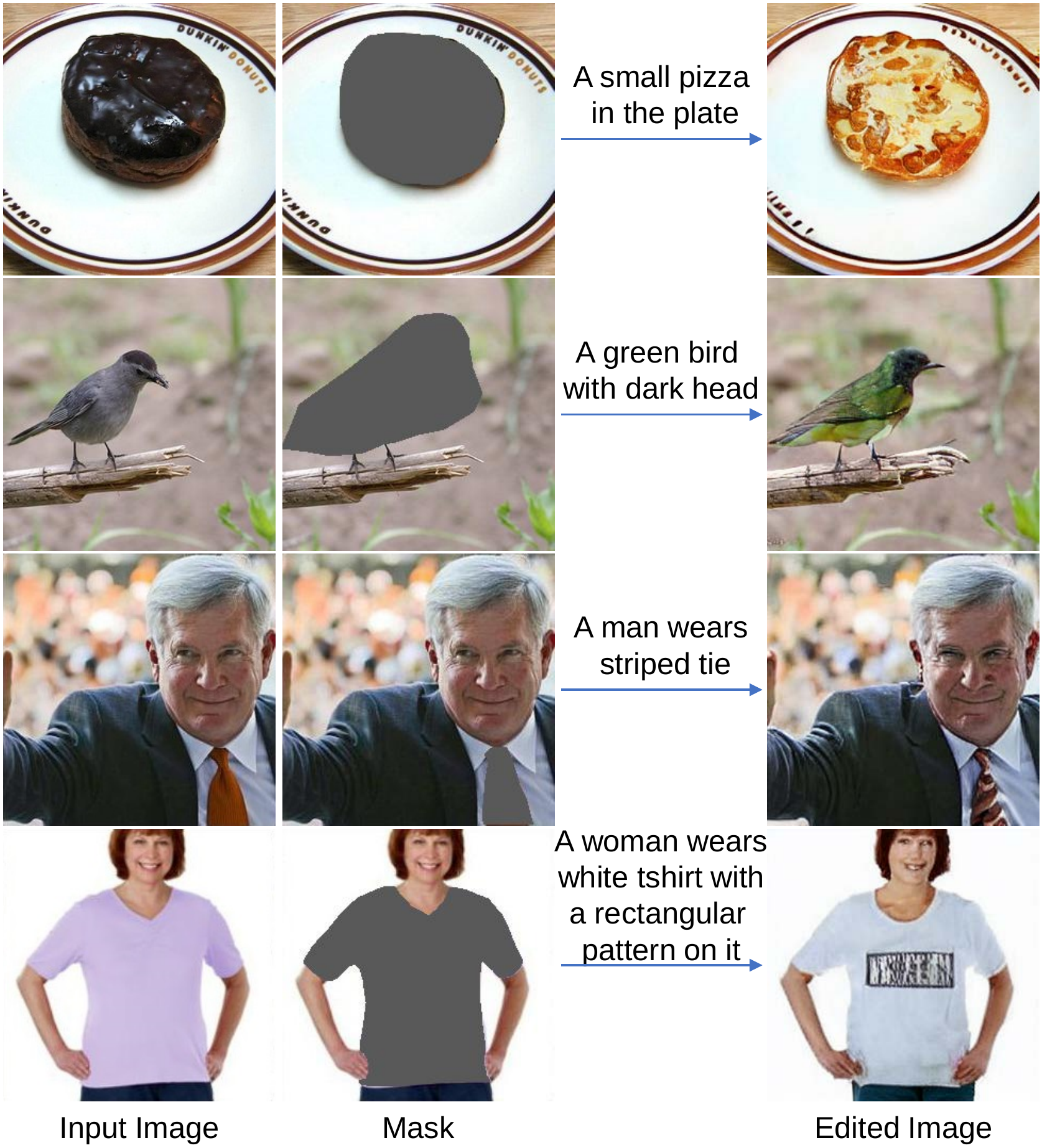}
	\vspace{-0.1cm}
	\caption{Text guided image editing by VQ-Diffusion.}
	\label{fig:editing}
	\vspace{-0.3cm}
\end{figure}

\begin{figure*}[t!]
	\centering
	\vspace{-0.4cm}
	\includegraphics[width=1.0\linewidth]{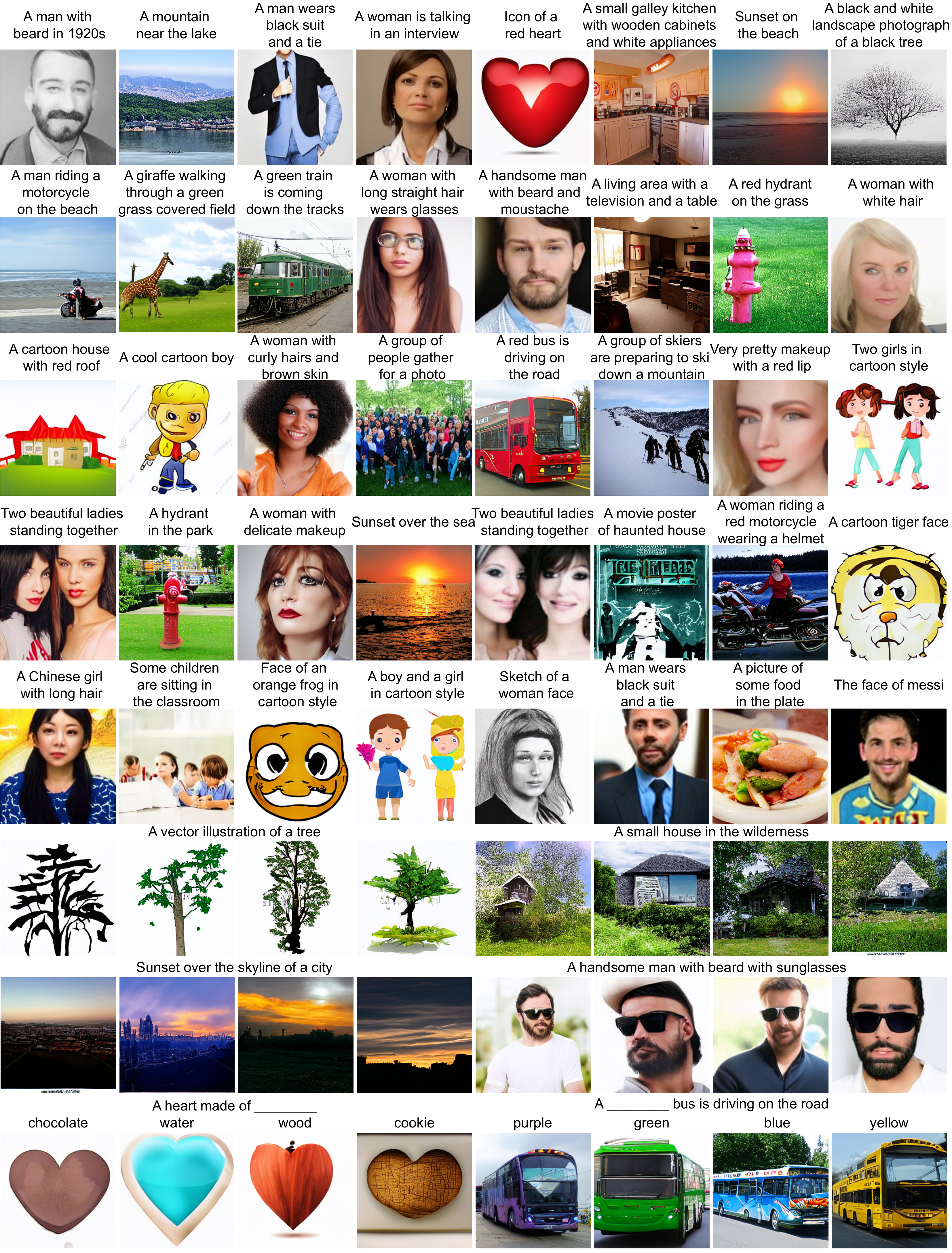}
	\vspace{-0.7cm}
	\caption{In the wild text-to-image synthesis by VQ-Diffusion.}
	\label{fig:result}
	\vspace{-0.3cm}
\end{figure*}

\begin{figure*}[t!]
	\centering
	\vspace{-0.4cm}
	\includegraphics[width=1.0\linewidth]{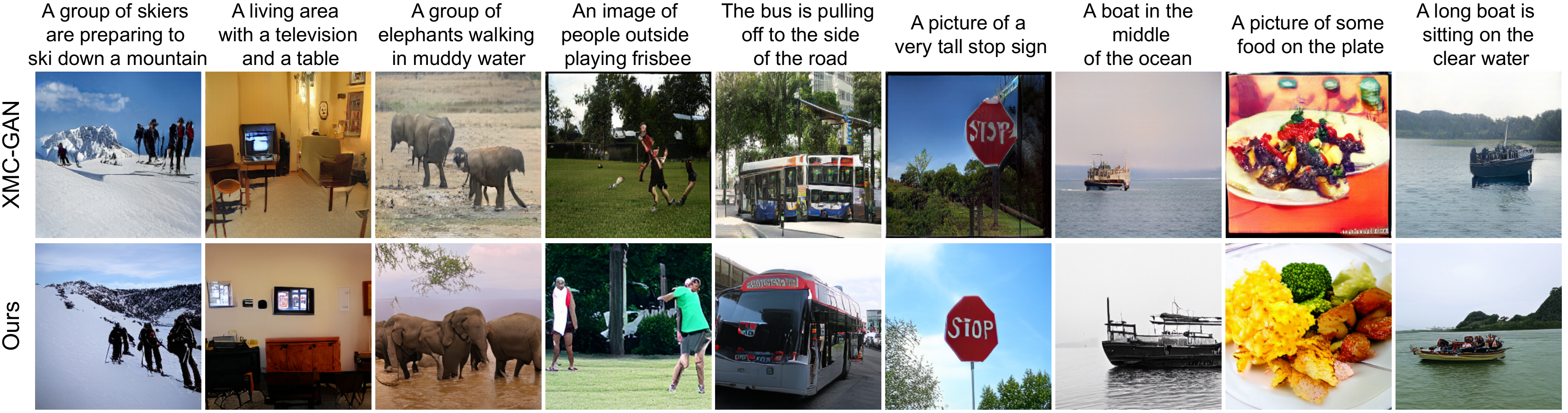}
	\vspace{-0.7cm}
	\caption{Comparison our results with XMC-GAN, their results come from their paper.}
	\label{fig:suppcom}
	\vspace{-0.3cm}
\end{figure*}

\begin{figure*}[t!]
	\centering
	\vspace{-0.4cm}
	\includegraphics[width=1.0\linewidth]{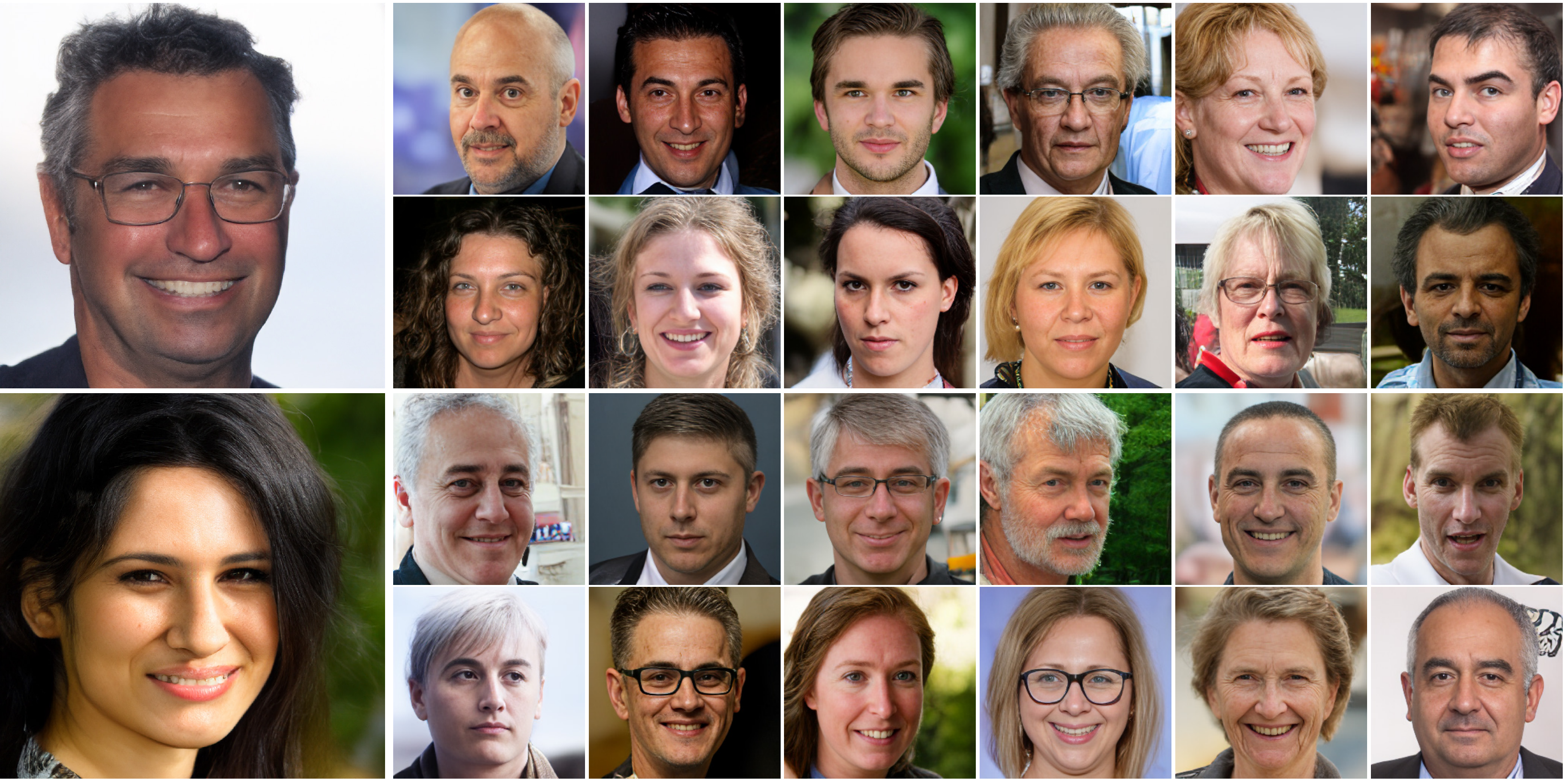}
	\vspace{-0.7cm}
	\caption{VQ-Diffusion results on FFHQ1024 and FFHQ256 datasets.}
	\label{fig:ffhq}
	\vspace{-0.3cm}
\end{figure*}

\begin{figure*}[t!]
	\centering
	\vspace{-0.4cm}
	\includegraphics[width=1.0\linewidth]{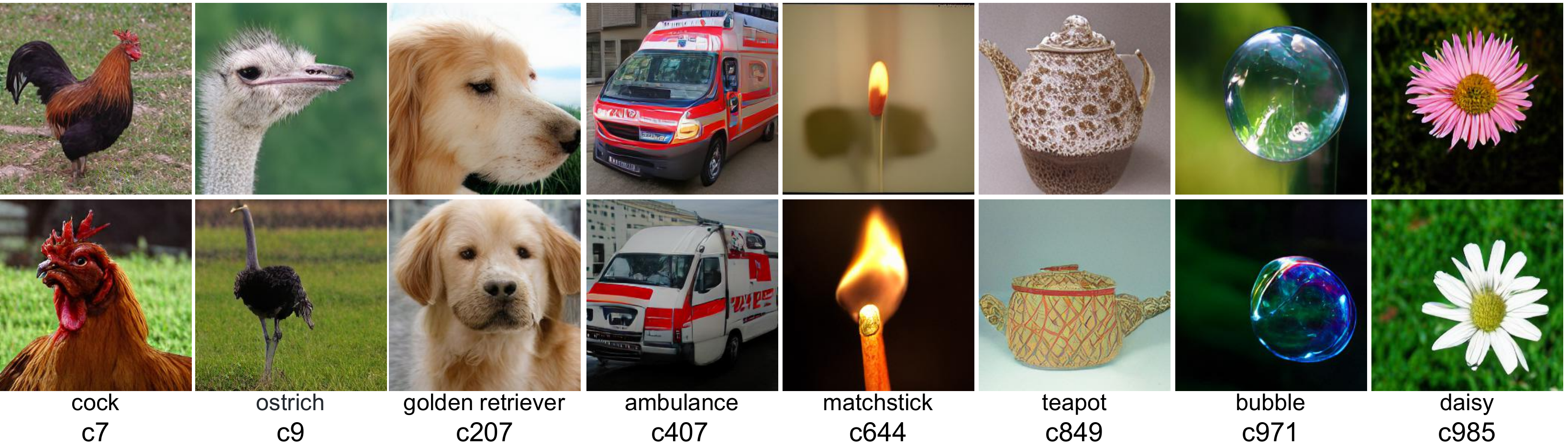}
	\vspace{-0.7cm}
	\caption{VQ-Diffusion results of class conditional synthesis on ImageNet dataset.}
	\label{fig:imagenet}
	\vspace{-0.3cm}
\end{figure*}

\section{Proof of Equation 8}

Mathematical induction can be used to prove the Equation 8 in the paper.

\noindent When $t=1$, we have
\begin{equation}
\label{eq_star}
    	\overline{\bm{Q}}_1 \bm{v}(x_0) = \begin{cases}
    	\overline{\alpha}_1 + \overline{\beta}_1, &x=x_0 \\
    	\overline{\beta}_1, & x \neq x_0 \,\text{and}\, x \neq K+1 \\
    	\overline{\gamma}_1, & x = K+1
    	\end{cases}
\end{equation}
which is clearly hold. Suppose the Equation 8 is hold at step $t$, then for $t=t+1$:
\begin{equation*}
    \overline{\bm{Q}}_{t+1} \bm{v}(x_0) = \bm{Q}_{t+1}\overline{\bm{Q}}_t \bm{v}(x_0)
\end{equation*}
When $x=x_0$,
\begin{equation*}
\small
\begin{split}
{\bm{Q}}_{t+1} \bm{v}(x_0)_{(x)} & = \overline{\beta}_t\beta_{t+1}(K-1) + (\alpha_{t+1}+\beta_{t+1})(\overline{\alpha}_t+\overline{\beta}_t) \\
& = \overline{\beta}_t(K\beta_{t+1}+\alpha_{t+1}) + \overline{\alpha}_t(\alpha_{t+1}+\beta_{t+1}) \\
    & =\frac{1}{K}(\overline{\beta}_t(1-\gamma_{t+1})+\overline{\alpha}_t\beta_{t+1} - \overline{\beta}_{t+1}) * K + \backslash \\
    &   \overline{\alpha}_{t+1} + \overline{\beta}_{t+1} \\
    & = \frac{1}{K}[(1-\overline{\alpha}_t-\overline{\gamma}_t)(1-\gamma_{t+1})+K\overline{\alpha}_t\beta_{t+1}- \backslash \\ 
    &  (1-\overline{\alpha}_{t+1}-\overline{\gamma}_{t+1})]+ \overline{\alpha}_{t+1} + \overline{\beta}_{t+1} \\
    & = \frac{1}{K}[(1-\overline{\gamma}_{t+1}) - \overline{\alpha}_t(1-\gamma_{t+1}-K\beta_{t+1})- \backslash \\ 
    & (1-\overline{\gamma}_{t+1})+  \overline{\alpha}_{t+1}] + \overline{\alpha}_{t+1} + \overline{\beta}_{t+1} \\
    & = \overline{\alpha}_{t+1} + \overline{\beta}_{t+1}
\end{split}
\end{equation*}
When $x=K+1$,
\begin{equation*}
\small
\begin{split}
{\bm{Q}}_{t+1} \bm{v}(x_0)_{(x)} & = \overline{\gamma}_t + (1-\overline{\gamma}_t)\gamma_{t+1} \\
 & = 1 - (1-\overline{\gamma}_{t+1}) \\
 & = \overline{\gamma}_{t+1}
\end{split}
\end{equation*}
When $x\neq x_0$ and $x \neq K+1$,
\begin{equation*}
\small
\begin{split}
    {\bm{Q}}_{t+1} \bm{v}(x_0)_{(x)}& =\overline{\beta}_t(\alpha_{t+1}+\beta_{t+1})+\overline{\beta}_t\beta_{t+1}(K-1) + \overline{\alpha}_t\beta_{t+1} \\
    & = \overline{\beta}_t(\alpha_{t+1}+\beta_{t+1} * K) + \overline{\alpha}_t\beta_{t+1} \\
    & = \frac{1-\overline{\alpha}_t-\overline{\gamma}_t}{K} * (1-\gamma_{t+1}) + \overline{\alpha}_t\beta_{t+1} \\ 
    & = \frac{1}{K}(1-\overline{\gamma}_{t+1}) + \overline{\alpha}_t (\beta_{t+1} - \frac{1-\gamma_{t+1}}{K}) \\
    & = \overline{\beta}_{t+1}
\end{split}
\end{equation*}

So proof done.

\section{Results}
In this part, we provide more visualization results. First, we compare our results with XMC-GAN in Figure~\ref{fig:suppcom}. We got their results directly from their paper. The irregular mask inpainting results are shown in Figure~\ref{fig:editing}. we show our more in the wild text-to-image results in Figure~\ref{fig:result}. And we provide our results on ImageNet and FFHQ in Figure~\ref{fig:imagenet} and Figure~\ref{fig:ffhq}.

\end{document}